\documentclass[onefignum,onetabnum]{siamart251216}

\usepackage{lipsum}
\usepackage{amsfonts}
\usepackage{graphicx}
\usepackage{epstopdf}
\usepackage{algorithmic}
\usepackage[numbers]{natbib}
\usepackage{xr-hyper}
\ifpdf
  \DeclareGraphicsExtensions{.eps,.pdf,.png,.jpg}
\else
  \DeclareGraphicsExtensions{.eps}
\fi

\newsiamremark{remark}{Remark}
\newsiamremark{hypothesis}{Hypothesis}
\crefname{hypothesis}{Hypothesis}{Hypotheses}
\newsiamthm{claim}{Claim}
\newsiamremark{fact}{Fact}
\crefname{fact}{Fact}{Facts}

\headers{Entropy-Regularized RL for Regime-Switching Models}{C. Hu, D. Li, X. Li, and Y. Zeng}

\title{Entropy-Regularized Reinforcement Learning for the Regime-Switching Consumption-Investment Problem\thanks{Submitted to the editors DATE.
\funding{This work was funded by XXXXXX.}}}

\author{Congde Hu\thanks{School of Statistics, East China Normal University, Shanghai, China 
  (\email{hcd.math@gmail.com}, \email{dpli@fem.ecnu.edu.cn}).}
\and Danping Li\footnotemark[2]\thanks{Corresponding author.}
\and Xun Li\thanks{Department of Applied Mathematics, The Hong Kong Polytechnic University, Kowloon, China 
  (\email{malixun@polyu.edu.hk}, \email{yi-jie.zeng@connect.polyu.hk}).}
\and Yijie Zeng\footnotemark[4]}

\usepackage{amsopn}

\ifpdf
\hypersetup{
  pdftitle={Entropy-Regularized Reinforcement Learning for the Regime-Switching Consumption-Investment Problem},
  pdfauthor={C. Hu, D. Li, X. Li, and Y. Zeng}
}
\fi

\begin{document}

\maketitle

\begin{abstract}
This paper establishes an entropy-regularized reinforcement learning (ERRL) framework for the continuous-time consumption-investment problem under a regime-switching market model with general utility functions. By incorporating randomized exploratory policies and Shannon entropy regularization into the stochastic optimal control formulation, we derive closed-form representations for the optimal exploratory consumption and investment strategies. Specifically, the optimal investment strategy follows a multivariate Gaussian distribution, whereas the optimal consumption strategy takes the form of a Gibbs measure, which reduces to a Gamma distribution under logarithmic utility. We also briefly illustrate the extension to power utility. Our analysis shows that the modes of the exploratory distributions recover the functional forms of the classical deterministic optimal controls, rather than generally coinciding with their means. Furthermore, under logarithmic utility, we establish the convergence of the exploratory value function to its classical unregularized counterpart as the temperature parameter vanishes. For algorithmic design, we develop a continuous-time $q$-learning framework grounded in a continuous-time $q$-function, prove the policy improvement theorem, and establish martingale characterizations for both on-policy and off-policy learning. We find that parameterizing the actor network based on theoretical distribution families can help accelerate convergence and enhance numerical stability. Furthermore, numerical experiments across regime-switching environments provide empirical support for our theoretical findings, demonstrating stable loss convergence as well as the monotonicity and concavity of the learned value functions.
\end{abstract}

\begin{keywords}
Stochastic control, Reinforcement learning, Consumption-investment, Regime-switching, General utility, q-learning
\end{keywords}

\begin{MSCcodes}
68Q25, 68R10, 68U05
\end{MSCcodes}

\section{Introduction}\label{sec:intro}
The continuous-time consumption-investment problem pioneered by \cite{merton1975optimum} provides a classical foundation for dynamic asset allocation under stochastic market environments. In Merton’s canonical model, investors maximize the lifetime expected utility derived from intermediate consumption and terminal wealth by solving a Hamilton–Jacobi–Bellman (HJB) equation, which admits closed-form deterministic optimal consumption and portfolio rules under logarithmic or constant relative risk aversion (CRRA) utilities. Recent work can be found in \cite{Bo2026}, \cite{Ma2022}, \cite{Zhou2026} and so on. In the financial market, investors usually face two market regimes, One represents a bull market characterized by rising asset prices, whereas the other represents a bear market characterized by declining asset prices. Therefore, the portfolio selection problem with regime-switching is of great interest and importance in financial investment. 
Under regime-switching models, optimal strategies for utility maximization have been extensively studied under various preferences, such as \cite{sotomayor2009explicit}, \cite{wang2026} and so on.

A practical limitation shared by classical stochastic optimal control approaches to consumption-investment problems is their full-information prerequisite: agents are assumed to perfectly observe and precisely estimate all market primitives, including Markov chain transition intensities, regime-dependent drifts and volatility matrices \cite{merton1980estimating}. In empirical applications, these structural parameters are often latent or subject to non-negligible estimation errors, which may render deterministic strategies less reliable or difficult to implement for real-time investment decisions. This observation motivates the integration of reinforcement learning (RL) with continuous-time stochastic control, which enables model-free policy learning through repeated environment interactions without requiring prior full knowledge of market dynamics.

Early continuous-time RL theories rooted in stochastic control were developed by \cite{wang2020reinforcement, wang2020continuous}, who introduced entropy-regularized exploratory policies for mean-variance portfolio selection and established a link between soft RL and HJB variational problems. Subsequent work provided martingale characterizations for temporal difference (TD) evaluation and continuous-time q-learning, offering a theoretical basis for actor-critic algorithms on state spaces \cite{jia2022policy,jia2023q,zhang2026continuous}. Extensions to alternative entropy regularizers have further illustrated the potential of exploratory stochastic control for financial tasks \cite{li2026,tang2022exploratory}. Despite these advances, the existing continuous-time RL literature has paid limited attention to the joint optimization of consumption and investment strategies within regime-switching markets, and to the statistical structure of optimal exploratory consumption and investment distributions under utility functions.

This paper develops an entropy-regularized reinforcement learning (ERRL) framework for the continuous-time consumption-investment optimization problem 
with general concave utility functions in a regime-switching diffusion market. We embed Shannon entropy regularization into the classical stochastic control objective to induce randomized exploratory policies, which relaxes the complete market information requirement of traditional Merton-type models. Our contributions are organized along theoretical, structural, and algorithmic-numerical dimensions.

First, we reformulate the regime-switching consumption-investment problem as an entropy-regularized stochastic optimal control problem with stochastic feedback policies, and derive the associated variational HJB equation for the regularized value function. Via calculus of variations over probability measures, the joint optimal exploratory action distribution can be separated into independent portfolio weight and consumption marginals: the optimal risky asset allocation follows a multivariate Gaussian law, while consumption obeys a Gibbs measure. An important contribution from our derivation reveals that the classical deterministic Merton optimal controls correspond to the \textit{mode} of the corresponding probability density, rather than the mean, of these exploratory distributions—a distinction that is particularly relevant for skewed consumption distributions such as the Gamma law under logarithmic utility. Under logarithmic utility, we also establish uniform convergence of the entropy-regularized value function to its classical unregularized counterpart as the temperature parameter vanishes, thereby connecting the exploratory RL framework with standard utility maximization theory.

Second, we investigate the optimal exploratory consumption policy under logarithmic utility and show that it follows a Gamma distribution with explicit shape and rate parameters. We further establish the convergence of the exploratory value function to its classical counterpart as the temperature parameter vanishes. As an additional illustration, we briefly discuss the corresponding Gibbs density under power utility and identify its mode. These results provide theoretical guidance for selecting suitable policy distributions in reinforcement learning algorithms.

Third, we construct and numerically validate a model-free continuous-time q-learning actor-critic algorithm adapted to regime-switching markets. Algorithmically, we introduce a regime-adjusted continuous-time q-function, establish a policy improvement result under entropy regularization, and derive martingale representations for both on-policy and off-policy learning pipelines. The proposed actor network employs a Gaussian distribution for investment and a Gamma-based parameterization for consumption, providing a tractable implementation of exploratory policies. To illustrate the applicability of our framework, we conduct numerical experiments on a two-state bull–bear simulator using logarithmic and power utilities. The results suggest stable training convergence and indicate that the learned value functions are generally consistent with economic properties, such as risk-averse concavity and a clear regime hierarchy (with bull-market values generally lying above bear-market values). Furthermore, the learned exploratory policies exhibit patterns consistent with the theoretical policy structures, reflecting intuitive consumption and investment shifts across alternating market conditions.

The structure of this paper is organized as follows. Section~\ref{Section 2} constructs the theoretical framework for the exploratory consumption-investment problem under the regime-switching model and derives the optimal exploratory policy. Section~\ref{Section 3} investigates the logarithmic utility case and briefly discusses its extension to power utility. Section~\ref{Section 4} designs a q-learning algorithm and provides numerical examples to illustrate our results. Finally, Section~\ref{Section 5} concludes the paper.

\textbf{Notation:} Throughout the paper, by convention, all vectors are column vectors unless otherwise specified.
Let $\mathbb R^N$ be the $N$-dimensional Euclidean space and $\mathbb R^{N\times N}$ be the space of real $N\times N$ matrices.
For a matrix $A$, $A^\top$ is its transpose, $A^{-1}$ its inverse,  $\det(A)$ its determinant, and $\Vert{}A\Vert{}_F$ its Frobenius norm. For a vector $x$, $\Vert{}x\Vert{}_2$ denotes its Euclidean norm. We use $I$ to denote the identity matrix and $\mathbf{1}$ to denote a vector of ones with appropriate dimensions.
For a sufficiently smooth function $v(t,x,i)$, $v_t$, $v_x$, and $v_{xx}$ denote the first-order partial derivative with respect to time $t$, and the first- and second-order partial derivatives with respect to the spatial variable $x$, respectively.
The symbol $\mathbb E_{t,x,i}[\cdot]$ denotes the conditional expectation given $X(t)=x$, $\alpha(t)=i$.
The symbol $\delta$ is used to denote the variational operator.
Let $\mathcal{A}$ denote the action space, and $\mathcal{P}(\mathcal{A})$ denote the set of probability density functions defined on $\mathcal{A}$. Finally, $\mathcal N(\cdot \mid\mu,\Sigma)$ denotes a multivariate Gaussian distribution with mean vector $\mu$ and covariance matrix $\Sigma$, and $\mathrm{Gamma}(\cdot \mid \alpha, \beta)$ denotes a Gamma distribution with shape parameter $\alpha$ and rate parameter $\beta$.

\section{Model formulation}\label{Section 2}
This section presents a mathematical formulation for the stochastic optimal control problem of exploratory consumption-investment decision. To provide a baseline for comparison, we first review the classical consumption-investment problem, following the setup in \cite{heunis2015utility}, and then introduce the exploratory consumption-investment problem by incorporating policy exploration into the decision process. Finally, we discuss the theoretical characterization of the proposed problem and derive the explicit form of the optimal  policy.
\subsection{Classical consumption-investment problem}\label{subsec:classical_model}
Let $(\Omega, \mathcal{F}, \{\mathcal{F}_t\}_{t \ge 0}, \mathbb{P})$ be a complete probability space. We consider a continuous-time Markov chain $\alpha(t)$ taking values in a finite state space $\mathcal{M} = \{1, 2, \dots, D\}$. The generator matrix of $\alpha(t)$ is given by $Q = (q_{ij})_{D \times D}$, such that for $i \neq j$, $q_{ij} \ge 0$ represents the transition rate from state $i$ to state $j$, and $q_{ii} = - \sum_{j \neq i} q_{ij}$. 
The market consists of one risk-free asset (bank account) and $N$ risky assets (stocks) traded continuously over a finite time horizon $[0, T]$. The market parameters depend on the current regime state $i \in \mathcal{M}$. For state $i \in \mathcal{M}$, let $r(t, i)$ be the risk-free interest rate, $b(t, i) \equiv (b_1(t, i), \dots, b_N(t, i))^\top \in \mathbb{R}^N$ be the vector of mean rates of return, and $\sigma(t, i) \equiv (\sigma_{nm}(t, i))_{N \times N} \in \mathbb{R}^{N \times N}$ be the volatility matrix. We assume the standard $N$-dimensional Brownian motion $W(t) = (W_1(t), \dots, W_N(t))^\top$ is independent of the Markov chain $\alpha(t)$. Furthermore, we assume that the filtration $\{\mathcal{F}_t\}_{t \ge 0}$ is generated by the Brownian motion $W(t)$ and the Markov chain $\alpha(t)$, augmented by all $\mathbb{P}$-null sets to satisfy the usual conditions.

The price process of the risk-free asset, denoted by $S_0(t)$, evolves according to the following ordinary differential equation (ODE),
\begin{equation}
    dS_0(t) = r(t, \alpha(t)) S_0(t) dt, \quad S_0(0) = s_0 > 0.
\end{equation}
The price processes of the $N$ risky assets, denoted by $S_n(t)$ for $n = 1, 2, \dots, N$, are governed by the following stochastic differential equations (SDEs),
\begin{equation}
    dS_n(t) = S_n(t) \Bigg[ b_n(t, \alpha(t)) dt + \sum_{m=1}^N \sigma_{nm}(t, \alpha(t)) dW_m(t) \Bigg], \quad S_n(0) = s_n > 0.
\end{equation}
For technical regularity, we assume that all the functions $r(t, i)$, $b_n(t, i)$, and $\sigma_{nm}(t, i)$ are measurable and uniformly bounded in $t$. Furthermore, we assume that the following nondegeneracy condition is satisfied for some $\delta > 0$,
\begin{equation}
    \sigma(t, i)\sigma^\top(t, i) \ge \delta I, \quad \forall t \in [0, T], \quad \forall i \in \mathcal{M},
\end{equation}
where $I$ is the $N \times N$ identity matrix. 

Let $u(t) \equiv (u_1(t), u_2(t), \dots, u_N(t))^\top \in \mathbb{R}^N$ denote the portfolio weight vector (the fraction of wealth invested in the $N$ risky assets), where negative components indicate short selling and a sum exceeding $1$ implies leveraging by borrowing at the risk-free rate. Let $c(t) \ge 0$ denote the consumption rate process. 
The wealth process $X(\cdot)$ under strategy $(u, c)$ evolves according to the following SDE for $s \in [t, T]$,
\begin{equation}
\left\{\begin{array}{ll}
dX(s) = &\Big\{ X(s) \big[ r(s, \alpha(s)) + u^\top(s) (b(s, \alpha(s)) - r(s, \alpha(s))\mathbf{1}) \big] - c(s) \Big\} ds \\
    &+ X(s) u^\top(s) \sigma(s, \alpha(s)) dW(s),\\
    X(t) =& x > 0,\quad \alpha(t)=i.
\end{array} \right.
\end{equation}
where, $\mathbf{1}$ is an $N$-dimensional vector of ones. 
For simplicity of notation, we introduce the market price of risk vector $\theta(t, i) := \sigma(t, i)^{-1}(b(t, i) - r(t, i)\mathbf{1})$. The wealth equation can be rewritten as
\begin{equation}\label{classical SDE}
\left\{\begin{array}{ll}
dX(s) = &\Big\{ X(s) \big[ r(s, \alpha(s)) + u^\top(s)\sigma(s, \alpha(s))\theta(s, \alpha(s)) \big] - c(s) \Big\} ds\\
&+ X(s) u^\top(s) \sigma(s, \alpha(s)) dW(s),\\
    X(t) =& x > 0,\quad \alpha(t)=i.
\end{array} \right.
\end{equation}

The classical performance functional is defined as
\begin{equation}\label{classical performance functional}
    J^{cl}(t,x,i; u, c) = \mathbb{E}_{t,x, i} \left[ \int_t^T U_1(s, c(s)) \, ds + U_2(X(T)) \right],
\end{equation}
where $\mathbb{E}_{t,x, i}[\cdot] := \mathbb{E}[\cdot \mid X(t)=x, \alpha(t)=i]$. 
We assume that the ``consumption'' and ``investment'' utility functions, $U_1 : [0,T] \times (0,\infty) \rightarrow \mathbb{R}$ and $U_2 : (0,\infty) \rightarrow \mathbb{R}$ respectively, satisfy the following standard conditions:
\begin{itemize}
    \item[(i)] For each $t \in [0,T]$, the mappings $x \rightarrow U_1(t,x)$ and $x \rightarrow U_2(x)$ are strictly increasing, strictly concave, and of class $C^1$ on $(0,\infty)$ (with $x$-derivatives denoted by $U_1^{'}(t,x)$ and $U_2^{'}(x)$ for each $(t,x) \in [0,T] \times (0,\infty)$) such that
    \begin{equation*}
    \begin{aligned}
        \lim_{x \downarrow 0} U_1^{'}(t, x) &= +\infty, \quad \lim_{x \rightarrow \infty} U_1^{'}(t, x) = 0, \\
        \lim_{x \downarrow 0} U_2^{'}(x) &= +\infty, \quad \lim_{x \rightarrow \infty} U_2^{'}(x) = 0;
    \end{aligned}
    \end{equation*}
    \item[(ii)] The mappings $(t,x) \rightarrow U_1(t,x)$ and $(t,x) \rightarrow U_1^{'}(t,x)$ are continuous on $[0,T] \times (0,\infty)$.
\end{itemize}

\textbf{Motivation for the exploratory framework.} In the classical framework \eqref{classical SDE}-\eqref{classical performance functional}, the standard stochastic optimal control problem is to find an optimal admissible strategy $(u^*, c^*)$ that maximizes the performance functional. While the classical regime-switching model provides a useful representation of financial reality compared to traditional models with deterministic coefficients, it generally assumes a known environment (\cite{merton1980estimating}). In this classical setting, the investor typically requires prior knowledge of the Markov chain's transition rates, as well as the drift, volatility, and interest rate parameters for each market regime. In practice, however, these parameters can be challenging to estimate precisely and may remain partially unknown. 

This practical consideration motivates the transition to an entropy-regularized reinforcement learning (ERRL) framework. By replacing deterministic controls with stochastic policies and incorporating a Shannon entropy regularizer, the agent is naturally encouraged to explore the state-action space (\cite{wang2020reinforcement}, \cite{jia2023q}). This exploration allows the investor to dynamically learn effective consumption and investment strategies through interactions with the environment, thereby reducing the reliance on complete a priori knowledge of the underlying market dynamics.

\subsection{Exploratory consumption-investment problem}\label{subsec:exploratory_model}
In RL, exploration is achieved by substituting deterministic actions with probability distributions. In our model, the control variables are the consumption rate $c \in [0, \infty)$ and the risky asset weight vector $u \in \mathbb{R}^N$, yielding the joint action space $\mathcal{A} := [0, \infty) \times \mathbb{R}^N$.

To enable learning in an unknown environment, let $\boldsymbol{\pi}: (t, x, i) \in [0, T] \times(0, \infty) \times \mathcal{M} \to \boldsymbol{\pi}(\cdot \mid t, x, i) \in \mathcal{P}(\mathcal{A})$ be a given stochastic feedback control, where $\mathcal{P}(\mathcal{A})$ is the set of probability density functions defined on the action space $\mathcal{A}$. The investor selects their consumption and investment strategy at time $s$ by sampling from the probability distribution $\boldsymbol{\pi}(c,u\mid s,X(s),\alpha(s))$.

Following the theoretical framework for continuous-time exploratory diffusions proposed by \cite{wang2020reinforcement} and further extended by \cite{gao2026reinforcement}, we reformulate the classical problem using stochastic strategies. The detailed derivation of this exploratory state dynamics is provided in Appendix \ref{Appendix A}. Under this exploratory framework, the state dynamics are governed by the corresponding regime-switching exploratory SDE
\begin{equation}\label{eq:exploratory_sde}
\left\{\begin{array}{ll}
dX(s) = &\displaystyle \bigg( \int_{\mathcal{A}} \Big\{ X(s) \big[ r(s, \alpha(s)) + u^\top \sigma(s, \alpha(s))\theta(s, \alpha(s)) \big] - c \Big\} \\
&\quad\quad\times\boldsymbol{\pi}(c, u \mid s, X(s), \alpha(s)) \,dc\,du \bigg) ds \\
&\displaystyle + \bigg\{ \int_{\mathcal{A}} X(s)^2 u^\top \sigma(s, \alpha(s))\sigma^\top(s, \alpha(s)) u\\
&\quad\quad\times\, \boldsymbol{\pi}(c, u \mid s, X(s), \alpha(s)) \,dc\,du \bigg\}^{1/2} d\tilde{W}(s), \\
X(t) = &x > 0, \quad \alpha(t) = i.
\end{array} \right.
\end{equation}
where $\tilde{W}(s)$ is a standard one-dimensional Brownian motion.

To encourage exploration, we add an entropy regularizer to the objective. The exploratory performance functional is defined as
\begin{equation}\label{eq:exploratory performance functional}
\begin{aligned}
J(t,x, i; \boldsymbol{\pi})
&= \mathbb{E}_{t,x,i}\bigg[ \int_t^T \Big( \int_{\mathcal{A}} U_1(s, c) \,\boldsymbol{\pi}(c, u \mid s, X(s), \alpha(s)) \,dc\,du \\
 - \lambda& \int_{\mathcal{A}} \boldsymbol{\pi}(c, u \mid s, X(s), \alpha(s)) \ln \boldsymbol{\pi}(c, u \mid s, X(s), \alpha(s)) \,dc\,du \Big) ds + U_2(X(T)) \bigg],
\end{aligned}
\end{equation}
where $\lambda > 0$ is the \textit{temperature parameter}.

\begin{remark}
In ERRL, entropy encourages exploration by increasing policy randomness. Since the investor aims to maximize their expected performance in the proposed framework, the positive entropy term $-\lambda \int_{\mathcal{A}} \boldsymbol{\pi} \ln \boldsymbol{\pi} \,dc\,du$ acts as an exploration reward. The constant $\lambda > 0$ serves as the \textit{temperature parameter}, influencing how quickly the investor updates their policy using new information and balancing the trade-off between exploitation (utility maximization) and the entropy incentive.
\end{remark}
\begin{definition}\label{def:admissible_strategy}
(Admissible policy).
The policy $\boldsymbol{\pi} = \boldsymbol{\pi}(\cdot \mid t, x, i)$ is called admissible, and we denote the set of admissible policies by $\boldsymbol{\Pi}$, if:
\begin{itemize}
    \item[$(\romannumeral 1)$] For all $(t, x, i) \in [0, T] \times (0, \infty) \times \mathcal{M}$, $\boldsymbol{\pi}(\cdot \mid t, x, i) \in \mathcal{P}(\mathcal{A})$.
    Moreover, for every Borel set $B \in \mathcal{B}(\mathcal{A})$, the mapping
    $
    (t, x, i) \longmapsto \int_B \boldsymbol{\pi}(c, u \mid t, x, i) \, dc\,du
    $
    is jointly measurable with respect to $\mathcal{B}([0,T]) \otimes \mathcal{B}((0,\infty)) \otimes 2^{\mathcal{M}}$.
    Consequently, for any $\{\mathcal{F}_s\}$-progressively measurable state process $(X(s), \alpha(s))$, the integrated drift and diffusion coefficients are $\{\mathcal{F}_s\}$-progressively measurable.
    \item[$(\romannumeral 2)$] For any constants $0 < \varepsilon < K < \infty$, all $t \in [0, T]$, and all $i \in \mathcal{M}$, there exists a constant $C_{\varepsilon,K} > 0$ such that for all $x, y \in [\varepsilon, K]$,
    \[
    \int_{\mathcal{A}} \left(1 + c + u^\top u\right) \left| \boldsymbol{\pi}(c,u \mid t, x, i) - \boldsymbol{\pi}(c, u \mid t, y, i) \right| dc\,du \le C_{\varepsilon,K} |x - y|.
    \]
    \item[$(\romannumeral 3)$] For all $(t, x, i) \in [0, T] \times (0, \infty) \times \mathcal{M}$ and some $p \ge 2$, there exists a constant $C > 0$ such that
    \[
    \int_{\mathcal{A}} \left| \ln \boldsymbol{\pi}(c, u \mid t, x, i) \right| \boldsymbol{\pi}(c, u \mid t, x, i) \,dc\,du \le C \left(1 +|\ln x|+ |x|^p\right).
    \]
    \item[$(\romannumeral 4)$] There exist constants $C > 0$ such that for all $(t, x, i) \in [0, T] \times (0, \infty) \times \mathcal{M}$,
    \[
    \int_{\mathcal{A}} c\, \boldsymbol{\pi}(c, u \mid t, x, i) \,dc\,du \le C x,\quad
    \int_{\mathcal{A}} u^\top u \, \boldsymbol{\pi}(c, u \mid t, x, i) \,dc\,du \le C.
    \]
    Furthermore, to ensure non-degeneracy of exploration, there exists a constant $\underline{\epsilon} > 0$ such that
    \[
    \int_{\mathcal{A}} u^\top u \, \boldsymbol{\pi}(c, u \mid t, x, i) \,dc\,du \ge \underline{\epsilon}, \quad \forall (t, x, i) \in [0, T] \times (0, \infty) \times \mathcal{M}.
    \]
\end{itemize}
\end{definition}

\begin{proposition}[Well-posedness of the exploratory SDE]\label{pro:Well-posedness of the exploratory SDE}
For any admissible policy $\boldsymbol{\pi} \in \boldsymbol{\Pi}$, the exploratory SDE \eqref{eq:exploratory_sde} admits a unique strong solution $\{X(s)\}_{s \in [t,T]}$ given the initial condition $X(t) = x > 0$ and $\alpha(t) = i$. Furthermore, the solution remains strictly positive almost surely, i.e., $\mathbb{P}(X(s) > 0, \forall s \in [t,T]) = 1$.
\end{proposition}
The proof can be found in Appendix \ref{Appendix B}.

\begin{definition}\label{def:value_function}
(Value function).
The exploratory value function is defined as
\begin{equation}\label{eq:value_function}
V(t, x, i) = \sup_{\boldsymbol{\pi} \in \boldsymbol{\Pi}} J(t, x, i; \boldsymbol{\pi}).
\end{equation}
\end{definition}

\subsection{Optimal exploratory  policy}\label{subsec:optimal_strategy}
In this subsection, we employ the Dynamic Programming Principle (DPP) to elucidate the optimal exploratory policy for the consumption-investment problem. First, we derive the HJB equation for the exploratory value function (see Lemma \ref{lem:hjb_exploratory}), and then present the optimal policy as a functional of the solution to the HJB equation (see Theorem \ref{thm:optimal_exploratory_policy}). Finally, we state a verification theorem (see Theorem \ref{thm:verification}) to guarantee that the solution to the HJB equation coincides with the true exploratory value function and that the generated policy is indeed optimal.

\begin{lemma}[HJB equation]\label{lem:hjb_exploratory}
Suppose that for all $i \in \mathcal{M}$, $v(\cdot,\cdot,i) \in C^{1,2}([0,T] \times (0, \infty))$. Then, the formal HJB equation associated with the exploratory consumption-investment problem is given by
\begin{equation}\label{eq:HJB_exploratory}
\begin{aligned}
-v_t(t,x,i) 
=&\sup_{\boldsymbol{\pi} \in \boldsymbol{\Pi}} \left[ \int_{\mathcal{A}} H(t, x, i, v_x, v_{xx}; c, u) \boldsymbol{\pi}(c, u \mid t, x, i) \,dc\,du \right.\\
&\left. -\lambda \int_{\mathcal{A}} \boldsymbol{\pi}(c, u \mid t, x, i) \ln \boldsymbol{\pi}(c, u \mid t, x, i) \,dc\,du \right] + \sum_{j=1}^{D} q_{ij} v(t,x,j),\\
v(T, x, i) = &U_2(x), \quad \forall x \in (0, \infty), \quad \forall i \in \mathcal{M},
\end{aligned}
\end{equation}
where the classical Hamiltonian $H$ is defined as
\begin{equation}\label{eq:classical_hamiltonian}
\begin{aligned}
H(t, x, i, v_x, v_{xx}; c, u) 
:=& U_1(t, c) + \left[ x \left( r(t, i) + u^\top \sigma(t, i)\theta(t, i) \right) - c \right] v_x\\
&+ \frac{1}{2} x^2 u^\top \sigma(t, i)\sigma^\top(t, i) u v_{xx} \\
=&  \left( \frac{1}{2} x^2 u^\top \sigma(t, i)\sigma^\top(t, i) u v_{xx} + x u^\top \sigma(t, i)\theta(t, i) v_x \right)\\
&+\left( U_1(t, c) - c v_x \right) + x r(t, i) v_x.
\end{aligned}
\end{equation}
\end{lemma}
\begin{proof}
The proof follows the standard DPP for regime-switching diffusions \cite{mao2006stochastic}, extended to the exploratory framework with entropy regularization \cite{wang2020reinforcement}.
\end{proof}
\begin{lemma}[Monotonicity and concavity]\label{lemma:Monotonicity and concavity}
For each $i\in\mathcal M$, suppose
$v(\cdot,\cdot,i)\in C^{1,2}([0,T)\times(0,\infty))$ is a finite-valued
 solution to the HJB equation \eqref{eq:HJB_exploratory}. Then
$v_x(t,x,i)>0$, $v_{xx}(t,x,i)<0$,
for all $(t,x,i)\in[0,T)\times(0,\infty)\times\mathcal M$.
\end{lemma}
The proof can be found in Appendix \ref{Appendix C}.

For completeness, we define the mode as follows \cite{ross2014first}.
\begin{definition}[Mode]\label{def:mode}
The mode of a continuous random variable having density $f$ is the value of $x$ for which $f(x)$ is a maximum.
\end{definition}

To solve the supremum problem in the HJB equation \eqref{eq:HJB_exploratory},
we employ the calculus of variations to maximize the objective functional with
respect to the probability density $\boldsymbol{\pi}$, subject to the normalization
constraint $\int_{\mathcal{A}} \boldsymbol{\pi}\,dc\,du = 1$.

\begin{theorem}[Optimal exploratory policy]
\label{thm:optimal_exploratory_policy}
Suppose that the HJB equation \eqref{eq:HJB_exploratory} admits a finite-valued solution $v(t,x,i)$. Then the optimal
exploratory policy $\boldsymbol{\pi}^*(c,u\mid t,x,i)$ separates into two
independent distributions,
$\boldsymbol{\pi}_c^*(c\mid t,x,i)$ and
$\boldsymbol{\pi}_u^*(u\mid t,x,i)$.
Specifically, the optimal investment policy is the multivariate Gaussian
distribution
\begin{equation}\label{eq:optimal_pi_a2}
\boldsymbol{\pi}_u^*(u \mid t, x, i)
=
\mathcal{N}\left(
u \,\middle|\,
-\frac{\sigma(t,i)^{-\top}\theta(t,i)v_x}{xv_{xx}},
\frac{\lambda(\sigma(t,i)\sigma^\top(t,i))^{-1}}
{-x^2v_{xx}}
\right),
\end{equation}
and the optimal consumption policy is the Gibbs distribution
\begin{equation}\label{eq:optimal_pi_a1}
\boldsymbol{\pi}_c^*(c \mid t, x, i)
=
\frac{
\exp\left(
\frac{1}{\lambda}[U_1(t,c)-cv_x]
\right)}
{
\int_0^\infty
\exp\left(
\frac{1}{\lambda}[U_1(t,c)-cv_x]
\right)dc
}.
\end{equation}
\end{theorem}
\begin{proof}
By Lemma~\ref{lemma:Monotonicity and concavity}, we have
$v_x>0$ and $v_{xx}<0$.

Starting from the HJB equation \eqref{eq:HJB_exploratory}, the maximizer
$\boldsymbol{\pi}^*(c,u\mid t,x,i)$ solves
\[
\sup_{\boldsymbol{\pi}\in\boldsymbol{\Pi}}
\left\{
\int_{\mathcal A}
H(t,x,i,v_x,v_{xx};c,u)\boldsymbol{\pi}\,dc\,du
-
\lambda\int_{\mathcal A}
\boldsymbol{\pi}\ln\boldsymbol{\pi}\,dc\,du
\right\},
\]
subject to $\int_{\mathcal A}\boldsymbol{\pi}\,dc\,du=1$.
By the method of Lagrange multipliers,
\[
\boldsymbol{\pi}^*(c,u\mid t,x,i)
=
\frac{
\exp\left(
\frac{1}{\lambda}
H(t,x,i,v_x,v_{xx};c,u)
\right)}
{
\int_{\mathcal A}
\exp\left(
\frac{1}{\lambda}
H(t,x,i,v_x,v_{xx};c,u)
\right)\,dc\,du
}.
\]

Using \eqref{eq:classical_hamiltonian}, the Hamiltonian can be decomposed as
\[
H
=
xr(t,i)v_x
+
\bigl(U_1(t,c)-cv_x\bigr)
+
\frac12 x^2 v_{xx}
u^\top\sigma(t,i)\sigma^\top(t,i)u
+
xv_xu^\top\sigma(t,i)\theta(t,i).
\]
Hence the joint density factorizes as
\[
\boldsymbol{\pi}^*(c,u\mid t,x,i)
=
\boldsymbol{\pi}_c^*(c\mid t,x,i)
\boldsymbol{\pi}_u^*(u\mid t,x,i).
\]

For the consumption policy,
\[
\boldsymbol{\pi}_c^*(c\mid t,x,i)
=
\frac{
\exp\left(
\frac{1}{\lambda}[U_1(t,c)-cv_x]
\right)}
{
\int_0^\infty
\exp\left(
\frac{1}{\lambda}[U_1(t,c)-cv_x]
\right)\,dc
}.
\]
Since $v_x>0$, the denominator is finite.

For the investment policy,
\[
\boldsymbol{\pi}_u^*(u\mid t,x,i)
\propto
\exp\left(
\frac{x^2v_{xx}}{2\lambda}
u^\top\sigma(t,i)\sigma^\top(t,i)u
+
\frac{xv_x}{\lambda}
u^\top\sigma(t,i)\theta(t,i)
\right).
\]
Since $v_{xx}<0$ and $\sigma(t,i)\sigma^\top(t,i)$ is positive definite,
the quadratic term is strictly negative definite. Completing the square gives
\[
\boldsymbol{\pi}_u^*(u \mid t, x, i)
=
\mathcal{N}\left(
u \,\middle|\,
-\frac{\sigma(t,i)^{-\top}\theta(t,i)v_x}{xv_{xx}},
\frac{\lambda(\sigma(t,i)\sigma^\top(t,i))^{-1}}
{-x^2v_{xx}}
\right).
\]
\end{proof}
\begin{remark}
The modes of the probability density functions
$\boldsymbol{\pi}_u^*(u \mid t, x, i)$ and
$\boldsymbol{\pi}_c^*(c \mid t, x, i)$, as defined in
Definition \ref{def:mode}, coincide with the pointwise
maximizers of the classical Hamiltonian $H$ evaluated
at the derivatives of the exploratory value function $v$.
Specifically, these modes are given by
\begin{equation*}
u_{\mathrm{mode}}^*
= -\frac{\sigma(t,i)^{-\top}\theta(t,i)v_x}
{xv_{xx}},
\qquad
c_{\mathrm{mode}}^*
= \left(U_1'(t,\cdot)\right)^{-1}(v_x).
\end{equation*}
These expressions are formally identical to the optimal
deterministic controls in the classical framework.

While prior studies (see, e.g., \cite{wang2020continuous})
have observed that the mean of the Gaussian exploratory
distribution formally recovers the classical optimal
control, our analysis highlights the role of the mode
in characterizing the pointwise Hamiltonian maximizer.
This distinction is particularly relevant for asymmetric
distributions, where the mean and mode generally differ.
To illustrate this observation, Section~\ref{Section 3}
considers a Gamma distribution for exploratory consumption,
whose mode retains the functional form of the classical
optimal consumption rule, whereas its mean differs from
the mode.
\end{remark}

Substituting the optimal exploratory policy $\boldsymbol{\pi}^*(c, u \mid t, x, i)=\boldsymbol{\pi}_c^*(c \mid t, x, i)\boldsymbol{\pi}_u^*(u \mid t, x, i)$ from Theorem \ref{thm:optimal_exploratory_policy} into the HJB equation \eqref{eq:HJB_exploratory}, we obtain
\begin{equation*}
\begin{aligned}
&\sup_{\boldsymbol{\pi} \in \boldsymbol{\Pi}}  \left[ \int_{\mathcal{A}} H(t, x, i, v_x, v_{xx}; c, u) \boldsymbol{\pi} \,dc\,du - \lambda \int_{\mathcal{A}} \boldsymbol{\pi} \ln \boldsymbol{\pi} \,dc\,du \right] \\
=& \lambda \ln \left( \int_{\mathcal{A}} \exp\left( \frac{H(t, x, i, v_x, v_{xx}; c, u)}{\lambda} \right) dc\,du \right) \\
=& x r(t, i) v_x- \frac{\theta(t, i)^\top \theta(t, i) v_x^2}{2 v_{xx}} - \frac{\lambda N}{2} \ln(-x^2 v_{xx}) - \frac{\lambda}{2} \ln\det(\sigma(t, i)\sigma^\top(t, i))\\
& + \frac{\lambda N}{2} \ln(2\pi\lambda)   + \lambda \ln \left( \int_0^\infty \exp\left( \frac{U_1(t, c) - c v_x}{\lambda} \right) dc \right).
\end{aligned}
\end{equation*}
Therefore, $v(t, x, i)$ satisfies the following  partial differential equation (PDE),
\begin{equation}\label{eq:explicit_HJB}
\begin{aligned}
-v_t(t, x, i) =& x r(t, i) v_x - \frac{\theta(t, i)^\top \theta(t, i) v_x^2}{2 v_{xx}} - \frac{\lambda N}{2} \ln(-x^2 v_{xx}) \\
&- \frac{\lambda}{2} \ln\det(\sigma(t, i)\sigma^\top(t, i))+ \frac{\lambda N}{2} \ln(2\pi\lambda) \\
&  + \lambda \ln \left( \int_0^\infty \exp\left( \frac{U_1(t, c) - c v_x}{\lambda} \right) dc \right) + \sum_{j=1}^{D} q_{ij} v(t,x,j),\\
v(T, x, i) = &U_2(x), \quad \forall x \in (0, \infty), \quad \forall i \in \mathcal{M}.
\end{aligned}
\end{equation}
\begin{theorem}[Verification theorem]\label{thm:verification}
Suppose that, for each $i\in\mathcal{M}$, there exists a finite-valued function
$v(t,x,i)\in C^{1,2}([0,T]\times(0,\infty))$ satisfying the following conditions:
\begin{itemize}
    \item[$(\romannumeral 1)$]
    $v$ solves the HJB equation \eqref{eq:HJB_exploratory} with terminal condition
    $v(T,x,i)=U_2(x)$;

    \item[$(\romannumeral 2)$]
    there exist constants $C>0$ and $p>0$ such that
    $
    |v(t,x,i)|
    \le C\left(1+x^p+x^{-p}\right)
    $
    for all $(t,x,i)\in[0,T]\times(0,\infty)\times\mathcal{M}$;

    \item[$(\romannumeral 3)$]
    for any admissible policy $\boldsymbol{\pi}\in\boldsymbol{\Pi}$,
    the local martingale term arising from the generalized It\^o formula applied to
    $v(s,X(s),\alpha(s))$ is a supermartingale. Furthermore, the optimal exploratory
    policy $\boldsymbol{\pi}^*$ defined in \eqref{eq:optimal_pi_a2} and
    \eqref{eq:optimal_pi_a1} is admissible, and the corresponding local martingale
    term is a true martingale.
\end{itemize}
Then
\[
v(t,x,i)
=
V(t,x,i)
=
\sup_{\boldsymbol{\pi}\in\boldsymbol{\Pi}}
J(t,x,i;\boldsymbol{\pi}),
\]
and $\boldsymbol{\pi}^*$ is an optimal exploratory policy.
\end{theorem}

\begin{proof}
By Lemma~\ref{lemma:Monotonicity and concavity},
$v_x>0$ and $v_{xx}<0$. The result then follows from the standard
verification argument for regime-switching stochastic control; see
\cite{sotomayor2009explicit,mao2006stochastic}.
\end{proof}

\section{Analysis under logarithmic utility}\label{Section 3}
In this section, we specialize the general optimal exploratory
consumption policy in Theorem~\ref{thm:optimal_exploratory_policy}
to logarithmic utility. We derive the resulting Gamma distribution
and analyze the convergence of the exploratory value function
as the temperature parameter tends to zero. We also briefly
discuss the extension to power utility.

Beyond their theoretical interest, the derived probability
density functions offer useful insights for practical algorithm
design. In particular, they provide a theoretical reference
for selecting parametric policy distributions in actor networks,
such as those used in soft actor-critic algorithms
\cite{haarnoja2018soft}. Choosing a distributional family
that reflects the underlying economic structure may facilitate
more stable and economically informed exploration in complex
continuous action spaces.

\begin{corollary}[Logarithmic utility]\label{cor:log_utility}
Suppose the investor has a logarithmic preference with a subjective discount rate
$\rho>0$, i.e.,
$
U_1(t,c)=e^{-\rho t}\ln(c),
$
and the HJB equation \eqref{eq:HJB_exploratory} admits a finite-valued 
solution $v(t,x,i)$. Then the optimal exploratory consumption policy is given by
\begin{equation}\label{eq:gamma_distribution}
\boldsymbol{\pi}_{c}^*(c \mid t,x,i)
=
\mathrm{Gamma}\left(
c \,\middle|\,
\frac{e^{-\rho t}}{\lambda}+1,\,
\frac{v_x(t,x,i)}{\lambda}
\right).
\end{equation}
\end{corollary}

We now establish the connection between the classical consumption-investment problem and the exploratory consumption-investment problem under logarithmic utility, by examining the asymptotic behavior of the model as the temperature parameter $\lambda \to 0^+$.
\begin{proposition}[Convergence of the value function under logarithmic utility]
\label{Thm_Value_Function_Convergence}
Suppose the investor has a logarithmic preference for consumption, i.e., $U_1(t, c) = e^{-\rho t} \ln(c)$. Let $V(t,x,i)$ denote the exploratory value function under temperature $\lambda > 0$, and $V^{cl}(t,x,i)$ denote the classical unregularized value function. As the exploration temperature $\lambda \to 0^+$, $V(t,x,i)$ converges uniformly to $V^{cl}(t,x,i)$ on any compact subset of the state space. Specifically, the approximation error is bounded by $\mathcal{O}(\lambda |\ln \lambda|)$.
\end{proposition}
The proof can be found in Appendix \ref{Appendix D}.

Building on the convergence of the value function established in Proposition \ref{Thm_Value_Function_Convergence}, we can now examine the connection between the derived exploratory policy $\boldsymbol{\pi}^*(c, u \mid t, x, i)$ and the classical deterministic  strategies. To do so, we analyze the asymptotic behavior of the exploratory policy distributions as the exploration temperature parameter $\lambda \to 0^+$.
It is worth noting that these consistency results rely on the uniform convergence of the exploratory value function $V(t,x,i)$ to its classical counterpart $V^{cl}(t,x,i)$ as $\lambda \to 0^+$. Across both policies, the modes of the respective probability density functions formally coincide with the classical optimal strategies, though their distributional properties exhibit distinct behaviors.
\begin{itemize}
\item Investment policy ($\boldsymbol{\pi}_u^*$): The optimal portfolio distribution follows a multivariate Gaussian, where the mode and the mean naturally coincide. As $\lambda \to 0^+$, the covariance matrix given by $-\frac{\lambda (\sigma(t, i)\sigma^\top(t, i))^{-1}}{x^2 V_{xx}}$ continuously shrinks toward the zero matrix. Consequently, the distribution collapses via a Dirac delta measure onto its mean vector, which precisely recovers the classical optimal portfolio proportion
\begin{equation*}
u^{*} = -\frac{\sigma(t, i)^{-\top}\theta(t, i) V_x}{x V_{xx}}.
\end{equation*}

\item Consumption policy ($\boldsymbol{\pi}_c^*$): Under logarithmic utility, the exploratory consumption policy follows a Gamma distribution with shape parameter $\alpha = \frac{e^{-\rho t}}{\lambda} + 1$ and rate parameter $\beta = \frac{V_x}{\lambda}$. For any $\lambda > 0$, the skewness introduced by entropy regularization causes the mode and the mean to differ. Specifically, while the expected value is $\mathbb{E}[c] = \frac{\alpha}{\beta} = \frac{e^{-\rho t} + \lambda}{V_x}$, the mode of the distribution precisely recovers the classical optimal consumption rate
\begin{equation*}
c^* = \frac{\alpha-1}{\beta} = \frac{e^{-\rho t}}{V_x}.
\end{equation*}
As $\lambda \to 0^+$, the variance
\begin{equation*}
\operatorname{Var}[c] = \frac{\lambda e^{-\rho t} + \lambda^2}{V_x^2}
\end{equation*}
vanishes. Consequently, the mean converges to the mode, and the distribution collapses via a Dirac delta measure centered at this classical deterministic consumption rate.
\end{itemize}

Together, these asymptotic properties demonstrate how the entropy-regularized exploratory framework smoothly transitions back to the classical framework as the exploration incentive fades away.

\begin{remark}[Power utility]\label{rem:power_utility}
The optimal exploratory consumption policy can also be characterized under power utility. Suppose that
\[
U_1(t,c)=e^{-\rho t}\frac{c^{1-\gamma}}{1-\gamma},
\qquad \gamma>0,\quad\gamma\neq1,
\]
and that the HJB equation \eqref{eq:HJB_exploratory} admits a finite-valued solution $v(t,x,i)$. By Theorem~\ref{thm:optimal_exploratory_policy}, the optimal consumption density is
\begin{equation}\label{eq:power_policy}
\boldsymbol{\pi}_{c}^*(c\mid t,x,i)
=\frac{
\displaystyle \exp\left(
\frac{e^{-\rho t}c^{1-\gamma}}{\lambda(1-\gamma)}
-\frac{v_x(t,x,i)}{\lambda}c
\right)
}{
\displaystyle \int_0^\infty\exp\left(
\frac{e^{-\rho t}z^{1-\gamma}}{\lambda(1-\gamma)}
-\frac{v_x(t,x,i)}{\lambda}z
\right)\,dz
},\qquad c>0.
\end{equation}
Unlike the logarithmic utility case, this density does not generally belong to a standard distribution family. Nevertheless, its mode is given explicitly by
\[
c_{\mathrm{mode}}^*(t,x,i)=
\bigl(e^{\rho t}v_x(t,x,i)\bigr)^{-1/\gamma},
\]
which has the same functional form as the classical optimal consumption rule, with the exploratory value function replacing its unregularized counterpart.
\end{remark}

\section{q-Learning RL algorithm}\label{Section 4}
While the optimal exploratory policy $\boldsymbol{\pi}^*(c, u \mid t, x, i)$ in Theorem \ref{thm:optimal_exploratory_policy} characterizes the exact solution to the exploratory optimal control problem, it relies on the market primitives $r(t, i)$, $b(t, i)$, $\sigma(t, i)$, and the transition rate matrix $Q$. In financial applications, these regime-switching dynamics are often difficult to estimate precisely or remain partially unknown a priori. To develop a model-free RL framework, we first define the associated action-value function ($q$-function) for the controller and establish its martingale characterization following continuous-time RL methodologies (cf. \cite{jia2022policy}, \cite{jia2023q}, \cite{zhang2026continuous}).

\subsection{$q$-Function and policy improvement}\label{sub41}
Based on the structure of the exploratory HJB equation \eqref{eq:HJB_exploratory}, we introduce the $q$-function, which evaluates the action-value of a deterministic action $(c, u) \in \mathcal{A}$ before the agent acts according to its stochastic policy.

\begin{definition}\label{definition q-function}
($q$-Function). For a given  $\boldsymbol{\pi}(\cdot \mid t, x, i)\in\boldsymbol{\Pi}$, let $J(t,x,i; \boldsymbol{\pi})$ be the corresponding exploratory payoff function. The $q$-function is defined as
\begin{equation}\label{q_function}
q(t, x, i; c, u; \boldsymbol{\pi}) := J_t(t, x, i; \boldsymbol{\pi}) + H(t, x, i, J_x,  J_{xx}; c, u) 
+ \sum_{j=1}^D q_{ij} J(t, x, j; \boldsymbol{\pi}),
\end{equation}
where $H$ is the classical Hamiltonian defined in \eqref{eq:classical_hamiltonian}.
\end{definition}
It immediately follows from the exploratory HJB equation that the $q$-function satisfies the following consistency condition
\begin{equation}\label{q_consistency}
\int_{\mathcal{A}} \left( q(t, x, i; c, u; \boldsymbol{\pi}) - \lambda \ln \boldsymbol{\pi}(c, u \mid t, x, i) \right) \boldsymbol{\pi}(c, u \mid t, x, i) \,dc\,du = 0, \quad \forall(t, x, i).
\end{equation}

\begin{lemma}\label{Policy improvement}
(Policy improvement).
Let $\boldsymbol{\pi}(c, u \mid t, x, i)\in \boldsymbol{\Pi}$ be an arbitrarily given admissible policy. Suppose that the corresponding exploratory performance function $J(\cdot,\cdot,i;\boldsymbol{\pi}) \in C^{1,2}([0,T) \times (0, \infty)) \cap C^{0}([0,T] \times (0, \infty))$ for each $i \in \mathcal{M}$, and satisfies $J_x > 0$ and $J_{xx} < 0$ (i.e., $J$ is strictly increasing and concave with respect to $x$). For any $(t,x,i)\in[0,T) \times (0, \infty) \times \mathcal{M}$, if the updated policy $\tilde{\boldsymbol{\pi}}(c, u \mid t,x,i)$ defined as
\begin{equation}\label{Policy improvement pi}
\tilde{\boldsymbol{\pi}}(c, u \mid t,x,i)=\frac{\exp \left(\frac{1}{\lambda}H(t,x,i,J_x,J_{xx};c, u)\right)}{\int_\mathcal{A} \exp \left(\frac{1}{\lambda}H(t,x,i,J_x,J_{xx};c', u')\right)dc'\,du'} 
\end{equation}
is admissible. Then, the exploratory performance functions under the specific policies $\tilde{\boldsymbol{\pi}}(c, u \mid t,x,i)$ and $\boldsymbol{\pi}(c, u \mid t,x,i)$ satisfy
\begin{equation}
J(t,x,i;\tilde{\boldsymbol{\pi}})\geq J(t,x,i;\boldsymbol{\pi}),\quad \forall (t,x,i)\in[0,T]\times(0, \infty) \times \mathcal{M}.
\end{equation}
\end{lemma}
\begin{proof}
Fix $(t, x, i)\in [0,T]\times(0, \infty) \times \mathcal{M}$. Let $X_s^{\tilde{\boldsymbol{\pi}}}$ denote the wealth process driven by the updated policy $\tilde{\boldsymbol{\pi}}(c, u \mid t, x, i)$. For notational simplicity, define the exploratory drift and diffusion coefficients under policy $\tilde{\boldsymbol{\pi}}$ as
\begin{align*}
b^{\tilde{\boldsymbol{\pi}}}(t, x, i) =& x r(t, i) + x \int_{\mathcal{A}} u^\top \sigma(t, i)\theta(t, i) \tilde{\boldsymbol{\pi}}(c, u \mid t, x, i) \,dc\,du \\
&- \int_{\mathcal{A}} c \tilde{\boldsymbol{\pi}}(c, u \mid t, x, i) \,dc\,du, \\
\sigma^{\tilde{\boldsymbol{\pi}}}(t, x, i) =& \left\{ x^2 \int_{\mathcal{A}} u^\top \sigma(t, i)\sigma^\top(t, i) u \, \tilde{\boldsymbol{\pi}}(c, u \mid t, x, i) \,dc\,du \right\}^{1/2}.
\end{align*}

Applying Itô's formula with regime switching to $J(s,X_{s}^{\tilde{\boldsymbol{\pi}}}, \alpha(s);\boldsymbol{\pi})$ along the path of $X_s^{\tilde{\boldsymbol{\pi}}}$, we have
\begin{equation*}
\begin{aligned}
&J(s,X_{s}^{\tilde{\boldsymbol{\pi}}}, \alpha(s);\boldsymbol{\pi})-J(t,x, i;\boldsymbol{\pi})\\
=
&\int^s_{t}\bigg[J_t(h,X_{h-}^{\tilde{\boldsymbol{\pi}}}, \alpha(h-);\boldsymbol{\pi}) + b^{\tilde{\boldsymbol{\pi}}}(h,X_{h-}^{\tilde{\boldsymbol{\pi}}}, \alpha(h-)) J_x(h,X_{h-}^{\tilde{\boldsymbol{\pi}}}, \alpha(h-);\boldsymbol{\pi})\\
&\,+\frac{1}{2} (\sigma^{\tilde{\boldsymbol{\pi}}}(h,X_{h-}^{\tilde{\boldsymbol{\pi}}}, \alpha(h-)))^2 J_{xx}(h,X_{h-}^{\tilde{\boldsymbol{\pi}}}, \alpha(h-);\boldsymbol{\pi})+\sum_{j=1}^D q_{\alpha(h-), j} J(h, X_{h-}^{\tilde{\boldsymbol{\pi}}}, j;\boldsymbol{\pi})\bigg]dh\\
&+\int^s_t J_x(h, X_{h-}^{\tilde{\boldsymbol{\pi}}}, \alpha(h-);\boldsymbol{\pi}) \sigma^{\tilde{\boldsymbol{\pi}}}(h,X_{h-}^{\tilde{\boldsymbol{\pi}}}, \alpha(h-))d\tilde{W}(h)+\int_t^s dM^{\alpha}_h ,
\end{aligned}
\end{equation*}
where $\tilde{W}$ is the effective 1-dimensional Brownian motion, and $M^{\alpha}_h$ is the zero-mean local martingale associated with jumps of the continuous-time Markov chain $\alpha$.

Define the localizing stopping times, for $n \geq 1$,
\begin{equation*}
\tau_n := \inf\left\{s \geq t : \int_t^s \left|J_x(h, X_{h-}^{\tilde{\boldsymbol{\pi}}}, \alpha(h-);\boldsymbol{\pi}) \cdot \sigma^{\tilde{\boldsymbol{\pi}}}(h,X_{h-}^{\tilde{\boldsymbol{\pi}}}, \alpha(h-))\right|^2dh \geq n \right\}.
\end{equation*} 
Taking the conditional expectation of both sides, the martingale terms vanish, and we obtain
\begin{equation}\label{eq:Ja}
\begin{aligned}   
&\mathbb{E}_{t,x,i}\left[ J(s\wedge\tau_n,X_{s\wedge\tau_n}^{\tilde{\boldsymbol{\pi}}}, \alpha(s\wedge\tau_n);\boldsymbol{\pi}) \right] - J(t,x,i;\boldsymbol{\pi})\\
=&\mathbb{E}_{t,x,i}\left[ \int^{s\wedge\tau_n}_{t}\bigg(J_t(h, X_{h-}^{\tilde{\boldsymbol{\pi}}}, \alpha(h-);\boldsymbol{\pi}) + b^{\tilde{\boldsymbol{\pi}}}(h,X_{h-}^{\tilde{\boldsymbol{\pi}}}, \alpha(h-)) J_x(h, X_{h-}^{\tilde{\boldsymbol{\pi}}}, \alpha(h-);\boldsymbol{\pi})\right.\\
+&\frac{1}{2} (\sigma^{\tilde{\boldsymbol{\pi}}}(h,X_{h-}^{\tilde{\boldsymbol{\pi}}}, \alpha(h-)))^2 J_{xx}(h, X_{h-}^{\tilde{\boldsymbol{\pi}}}, \alpha(h-);\boldsymbol{\pi})+\sum_{j=1}^D q_{\alpha(h-), j} J(h, X_{h-}^{\tilde{\boldsymbol{\pi}}}, j;\boldsymbol{\pi})\bigg)dh\bigg].
\end{aligned}
\end{equation}

On the other hand, by the Feynman-Kac equation for the performance function under fixed policy $\boldsymbol{\pi}$, we have the equality for all $(t,x,i)\in[0, T) \times (0, \infty) \times \mathcal{M}$,
\begin{equation}\label{eq:Ha_eq}
\begin{aligned}
-J_t(t,x,i;\boldsymbol{\pi})
=&\int_{\mathcal{A}}H(t, x, i, J_x, J_{xx};c, u)\,\boldsymbol{\pi}(c, u \mid t, x, i)\,dc\,du \\
&-\lambda\int_{\mathcal{A}}\boldsymbol{\pi}(c, u \mid t, x, i) \ln \boldsymbol{\pi}(c, u \mid t, x, i)\,dc\,du + \sum_{j=1}^{D} q_{ij} J(t, x, j;\boldsymbol{\pi}).
\end{aligned}
\end{equation}
By definition, $\tilde{\boldsymbol{\pi}}$ is the maximizer of the functional $\int_{\mathcal{A}} H \boldsymbol{\pi} \,dc\,du - \lambda \int_{\mathcal{A}} \boldsymbol{\pi}\ln\boldsymbol{\pi} \,dc\,du$ over all admissible policies. Thus for any policy $\boldsymbol{\pi}$,
\begin{equation*}
\int_{\mathcal{A}} H \boldsymbol{\pi} \,dc\,du - \lambda \int_{\mathcal{A}} \boldsymbol{\pi}\ln\boldsymbol{\pi} \,dc\,du \leq \int_{\mathcal{A}} H \tilde{\boldsymbol{\pi}} \,dc\,du - \lambda \int_{\mathcal{A}} \tilde{\boldsymbol{\pi}}\ln\tilde{\boldsymbol{\pi}} \,dc\,du.
\end{equation*}
Substituting this into \eqref{eq:Ha_eq} evaluated along $(h, X_{h-}^{\tilde{\boldsymbol{\pi}}}, \alpha(h-))$ yields
\begin{equation}\label{eq:Ha_ineq}
\begin{aligned}
&-J_t(h, X_{h-}^{\tilde{\boldsymbol{\pi}}}, \alpha(h-);\boldsymbol{\pi})
\\
\leq&\int_{\mathcal{A}}H(h, X_{h-}^{\tilde{\boldsymbol{\pi}}}, \alpha(h-), J_x, J_{xx};c, u)\,\tilde{\boldsymbol{\pi}}(c, u \mid h, X_{h-}^{\tilde{\boldsymbol{\pi}}}, \alpha(h-))\,dc\,du \\
&-\lambda\int_{\mathcal{A}}\tilde{\boldsymbol{\pi}}(c, u \mid h, X_{h-}^{\tilde{\boldsymbol{\pi}}}, \alpha(h-)) \ln \tilde{\boldsymbol{\pi}}(c, u \mid h, X_{h-}^{\tilde{\boldsymbol{\pi}}}, \alpha(h-))\,dc\,du \\
&+ \sum_{j=1}^{D} q_{\alpha(h-), j} J(h, X_{h-}^{\tilde{\boldsymbol{\pi}}}, j;\boldsymbol{\pi}).
\end{aligned}
\end{equation}

Recall the classical Hamiltonian definition \eqref{eq:classical_hamiltonian}:
$$
\int_{\mathcal{A}} H \tilde{\boldsymbol{\pi}} \,dc\,du = \int_{\mathcal{A}} U_1(h,c) \tilde{\boldsymbol{\pi}} \,dc\,du + b^{\tilde{\boldsymbol{\pi}}} J_x + \frac{1}{2}(\sigma^{\tilde{\boldsymbol{\pi}}})^2 J_{xx}.
$$
Substituting it into \eqref{eq:Ha_ineq} and re-arranging terms lead to 
\begin{equation*}
J_t + b^{\tilde{\boldsymbol{\pi}}} J_x + \frac{1}{2}(\sigma^{\tilde{\boldsymbol{\pi}}})^2 J_{xx} + \sum_{j=1}^D q_{\alpha(h-), j} J \geq -\int_{\mathcal{A}} U_1 \tilde{\boldsymbol{\pi}} \,dc\,du + \lambda \int_{\mathcal{A}} \tilde{\boldsymbol{\pi}} \ln \tilde{\boldsymbol{\pi}} \,dc\,du.
\end{equation*}
Plugging the above inequality into \eqref{eq:Ja}, we get
\begin{equation*}
\begin{aligned}
J(t,x,i;\boldsymbol{\pi})
\leq& \mathbb{E}_{t,x,i}\bigg[\int_t^{s\wedge\tau_n} \bigg(\int_{\mathcal{A}} U_1(h, c) \tilde{\boldsymbol{\pi}}(c, u \mid h, X_{h-}^{\tilde{\boldsymbol{\pi}}}, \alpha(h-)) dc\,du\\
&-\lambda\int_{\mathcal{A}} \tilde{\boldsymbol{\pi}}(c, u \mid h, X_{h-}^{\tilde{\boldsymbol{\pi}}}, \alpha(h-)) \ln \tilde{\boldsymbol{\pi}}(c, u \mid h, X_{h-}^{\tilde{\boldsymbol{\pi}}}, \alpha(h-))\, dc\,du \bigg)d h \\
&+J(s\wedge\tau_n,X^{\tilde{\boldsymbol{\pi}}}_{s\wedge\tau_n}, \alpha(s\wedge\tau_n);\boldsymbol{\pi})\bigg].
\end{aligned}
\end{equation*}

Set $s = T$ and let $n \to \infty$. By the dominated convergence theorem and the terminal condition $J(T, X_T^{\tilde{\boldsymbol{\pi}}}, \alpha(T);\boldsymbol{\pi})= U_2(X_T^{\tilde{\boldsymbol{\pi}}})$, we derive that for all $(t,x,i) \in [0,T] \times (0, \infty) \times \mathcal{M}$,
\[
\begin{aligned}
J(t,x,i;\boldsymbol{\pi})
&\leq \mathbb{E}_{t,x,i}\bigg[\int_t^{T} \bigg(\int_{\mathcal{A}} U_1(h, c) \tilde{\boldsymbol{\pi}}(c, u \mid h, X_{h-}^{\tilde{\boldsymbol{\pi}}}, \alpha(h-)) dc\,du\\
-\lambda\int_{\mathcal{A}}& \tilde{\boldsymbol{\pi}}(c, u \mid h, X_{h-}^{\tilde{\boldsymbol{\pi}}}, \alpha(h-)) \ln \tilde{\boldsymbol{\pi}}(c, u \mid h, X_{h-}^{\tilde{\boldsymbol{\pi}}}, \alpha(h-))\, dc\,du \bigg)d h + U_2(X_T^{\tilde{\boldsymbol{\pi}}})\bigg]
\\
&=J(t,x,i;\tilde{\boldsymbol{\pi}}).
\end{aligned}
\]
\end{proof}

\begin{theorem}\label{Policy improvement via Q-function}
(Policy improvement via $q$-function).
Let $\boldsymbol{\pi}(c, u \mid t, x, i)\in \boldsymbol{\Pi}$ be an arbitrarily given admissible policy. Suppose that the corresponding exploratory performance function $J(\cdot,\cdot,i;\boldsymbol{\pi}) \in C^{1,2}([0,T) \times (0, \infty)) \cap C^{0}([0,T] \times (0, \infty))$ for each $i \in \mathcal{M}$, and satisfies $J_x > 0$ and $J_{xx} < 0$ (i.e., $J$ is strictly increasing and concave with respect to $x$). For any $(t,x,i)\in[0,T) \times (0, \infty) \times \mathcal{M}$, if the updated policy $\tilde{\boldsymbol{\pi}}(c, u \mid t,x,i)$ defined as
\begin{equation}\label{q_str_improvement}
\tilde{\boldsymbol{\pi}}(c, u \mid t,x,i) = \frac{\exp \left( \frac{1}{\lambda} q(t, x, i; c, u; \boldsymbol{\pi}) \right)}{\int_{\mathcal{A}} \exp \left( \frac{1}{\lambda} q(t, x, i; c', u'; \boldsymbol{\pi}) \right) dc'\,du'}
\end{equation}
is admissible, then the exploratory performance function under the new policy satisfies
\begin{equation}
J(t,x,i; \tilde{\boldsymbol{\pi}}) \geq J(t,x,i; \boldsymbol{\pi}).
\end{equation}
\end{theorem}
\begin{proof}
The proof follows directly from Definition \ref{definition q-function} and Lemma \ref{Policy improvement}.
\end{proof}
\subsection{Martingale characterization of the $q$-function}\label{sub42}
Following the ideas in \cite{jia2023q,zhang2026continuous}, this subsection specializes the martingale characterization for the continuous-time exploratory state process within a regime-switching setting.

\begin{lemma}[Characterization of the exploratory $q$-function]\label{lemma:q_function_characterization}
Let an admissible policy $\boldsymbol{\pi} \in \boldsymbol{\Pi}$ be given, and suppose its value function $J(\cdot, \cdot, i; \boldsymbol{\pi}) \in C^{1,2}([0, T] \times (0, \infty)) \cap C([0, T] \times (0, \infty))$ for all $i \in \mathcal{M}$. Let a continuous function $\hat{q}: [0, T] \times (0, \infty) \times \mathcal{M} \times \mathcal{A} \rightarrow \mathbb{R}$ be given. Then, we have
\begin{itemize}
    \item[(i)] $\hat{q}(t, x, i, c, u) = q(t, x, i, c, u; \boldsymbol{\pi})$ for all $(t, x, i, c, u) \in [0, T] \times (0, \infty) \times \mathcal{M} \times \mathcal{A}$ if and only if for all $(t, x, i) \in [0, T] \times (0, \infty) \times \mathcal{M}$, the following process
    \begin{equation}\label{eq:martingale_condition_1}
    \begin{aligned}
        &J(s, X^{\boldsymbol{\pi}}(s), \alpha(s); \boldsymbol{\pi})\\
        &+ \int_t^s \left[ U_1(v, c^{\boldsymbol{\pi}}(v)) - \hat{q}(v, X^{\boldsymbol{\pi}}(v), \alpha(v), c^{\boldsymbol{\pi}}(v), u^{\boldsymbol{\pi}}(v)) \right] dv,
        \end{aligned}
    \end{equation}
    is an $(\{\mathcal{F}_s\}_{s \ge t}, \mathbb{P})$-martingale, where $\{X^{\boldsymbol{\pi}}(s), t \le s \le T\}$ is the solution to the exploratory SDE under $\boldsymbol{\pi}$ with initial conditions $X^{\boldsymbol{\pi}}(t) = x$, $\alpha(t) = i$, and $(c^{\boldsymbol{\pi}}(v), u^{\boldsymbol{\pi}}(v))$ is the action process sampled from $\boldsymbol{\pi}(\cdot \mid v, X^{\boldsymbol{\pi}}(v), \alpha(v))$.

    \item[(ii)] If $\hat{q}(t, x, i, c, u) = q(t, x, i, c, u; \boldsymbol{\pi})$ for all $(t, x, i, c, u) \in [0, T] \times (0, \infty) \times \mathcal{M} \times \mathcal{A}$, then given any other admissible policy $\boldsymbol{\pi}' \in \boldsymbol{\Pi}$, for all $(t, x, i) \in [0, T] \times (0, \infty) \times \mathcal{M}$, the following process
    \begin{equation}\label{eq:martingale_condition_2}
    \begin{aligned}
        &J(s, X^{\boldsymbol{\pi}'}(s), \alpha(s); \boldsymbol{\pi})\\
        &+ \int_t^s \left[ U_1(v, c^{\boldsymbol{\pi}'}(v)) - \hat{q}(v, X^{\boldsymbol{\pi}'}(v), \alpha(v), c^{\boldsymbol{\pi}'}(v), u^{\boldsymbol{\pi}'}(v)) \right] dv,
        \end{aligned}
    \end{equation}
    is an $(\{\mathcal{F}_s\}_{s \ge t}, \mathbb{P})$-martingale. Here, $\{X^{\boldsymbol{\pi}'}(s), t \le s \le T\}$ is the solution to the exploratory SDE under $\boldsymbol{\pi}'$ with $X^{\boldsymbol{\pi}'}(t) = x$, $\alpha(t) = i$, and $(c^{\boldsymbol{\pi}'}(v), u^{\boldsymbol{\pi}'}(v))$ is the action process sampled from $\boldsymbol{\pi}'(\cdot \mid v, X^{\boldsymbol{\pi}'}(v), \alpha(v))$.

    \item[(iii)] If there exists an admissible policy $\boldsymbol{\pi}' \in \boldsymbol{\Pi}$ such that for all $(t, x, i) \in [0, T] \times (0, \infty) \times \mathcal{M}$, the process defined in \eqref{eq:martingale_condition_2} is an $(\{\mathcal{F}_s\}_{s \ge t}, \mathbb{P})$-martingale with initial condition $X^{\boldsymbol{\pi}'}(t) = x, \alpha(t) = i$, then $\hat{q}(t, x, i, c, u) = q(t, x, i, c, u; \boldsymbol{\pi})$ for all $(t, x, i, c, u) \in [0, T] \times (0, \infty) \times \mathcal{M} \times \mathcal{A}$.
\end{itemize}
Moreover, in any of the three cases above, the $q$-function satisfies
\begin{equation}\label{eq:q_consistency_condition}
    \int_{\mathcal{A}} \left[ q(t, x, i, c, u; \boldsymbol{\pi}) - \lambda \ln \boldsymbol{\pi}(c, u \mid t, x, i) \right] \boldsymbol{\pi}(c, u \mid t, x, i) dc \, du = 0.
\end{equation}
\end{lemma}
The proof follows the same argument as that of Proposition 3.4
in \cite{zhang2026continuous} and is therefore omitted.

As in Theorem~3.5 of \cite{zhang2026continuous}, we can
strengthen Lemma~\ref{lemma:q_function_characterization}
to jointly characterize the q-function and the value
function associated with a given policy $\boldsymbol{\pi}$.
\begin{lemma}[Martingale characterization]\label{thm:Martingale Characterization Theorem}
Let an admissible policy $\boldsymbol{\pi} \in \boldsymbol{\Pi}$, a function $\hat{J}(\cdot, \cdot, i) \in C^{1,2}([0, T] \times (0, \infty)) \cap C([0, T] \times (0, \infty))$ for all $i \in \mathcal{M}$, and a continuous function $\hat{q}: [0, T] \times (0, \infty) \times \mathcal{M} \times \mathcal{A} \rightarrow \mathbb{R}$ be given such that, for all $(t, x, i, c, u) \in [0, T] \times (0, \infty) \times \mathcal{M} \times \mathcal{A}$,
\begin{equation}\label{eq:theorem_conditions}
\left\{\begin{array}{ll}
 \hat{J}(T, x, i) = U_2(x),\\
\int_{\mathcal{A}} \left[ \hat{q}(t, x, i, c, u) - \lambda \ln \boldsymbol{\pi}(c, u \mid t, x, i) \right] \boldsymbol{\pi}(c, u \mid t, x, i) \, dc \, du = 0.
\end{array} \right.
\end{equation}
Then,
\begin{itemize}
    \item[(i)] $\hat{J}(t, x, i)$ and $\hat{q}(t, x, i, c, u)$ are respectively the value function and the $q$ function associated with $\boldsymbol{\pi}$ if and only if for all $(t, x, i) \in [0, T] \times (0, \infty) \times \mathcal{M}$, the following process
    \begin{equation}\label{eq:thm_martingale_1}
        \hat{J}(s, X^{\boldsymbol{\pi}}(s), \alpha(s)) + \int_t^s \left[ U_1(v, c^{\boldsymbol{\pi}}(v)) - \hat{q}(v, X^{\boldsymbol{\pi}}(v), \alpha(v), c^{\boldsymbol{\pi}}(v), u^{\boldsymbol{\pi}}(v)) \right] dv,
    \end{equation}
    is an $(\{\mathcal{F}_s\}_{s \ge t}, \mathbb{P})$-martingale, where $\{X^{\boldsymbol{\pi}}(s), t \le s \le T\}$ is the solution to the exploratory SDE under $\boldsymbol{\pi}$ with initial conditions $X^{\boldsymbol{\pi}}(t) = x$, $\alpha(t) = i$, and $(c^{\boldsymbol{\pi}}(v), u^{\boldsymbol{\pi}}(v))$ is the action process sampled from $\boldsymbol{\pi}(\cdot \mid v, X^{\boldsymbol{\pi}}(v), \alpha(v))$.
    
    \item[(ii)] If $\hat{J}(t, x, i)$ and $\hat{q}(t, x, i, c, u)$ are respectively the value function and the $q$-function associated with $\boldsymbol{\pi}$, then for any admissible policy $\boldsymbol{\pi}' \in \boldsymbol{\Pi}$ and all $(t, x, i) \in [0, T] \times (0, \infty) \times \mathcal{M}$, the following process
    \begin{equation}\label{eq:thm_martingale_2}
        \hat{J}(s, X^{\boldsymbol{\pi}'}(s), \alpha(s)) + \int_t^s \left[ U_1(v, c^{\boldsymbol{\pi}'}(v)) - \hat{q}(v, X^{\boldsymbol{\pi}'}(v), \alpha(v), c^{\boldsymbol{\pi}'}(v), u^{\boldsymbol{\pi}'}(v)) \right] dv,
    \end{equation}
    is an $(\{\mathcal{F}_s\}_{s \ge t}, \mathbb{P})$-martingale, where $\{X^{\boldsymbol{\pi}'}(s), t \le s \le T\}$ is the solution to the exploratory SDE under $\boldsymbol{\pi}'$ with $X^{\boldsymbol{\pi}'}(t) = x, \alpha(t) = i$, and $(c^{\boldsymbol{\pi}'}(v), u^{\boldsymbol{\pi}'}(v))$ is the action process sampled from $\boldsymbol{\pi}'(\cdot \mid v, X^{\boldsymbol{\pi}'}(v), \alpha(v))$.
    
    \item[(iii)] If there exists $\boldsymbol{\pi}' \in \boldsymbol{\Pi}$ such that for all $(t, x, i) \in [0, T] \times (0, \infty) \times \mathcal{M}$, the process \eqref{eq:thm_martingale_2} is an $(\{\mathcal{F}_s\}_{s \ge t}, \mathbb{P})$-martingale with initial condition $X^{\boldsymbol{\pi}'}(t) = x, \alpha(t) = i$, then $\hat{J}(t, x, i)$ and $\hat{q}(t, x, i, c, u)$ are respectively the value function and the $q$-function associated with $\boldsymbol{\pi}$.
\end{itemize}
Moreover, in any of the three cases above, if it holds further that $\boldsymbol{\pi}$ takes the form of the Gibbs measure
\begin{equation}\label{eq:optimal_gibbs_policy}
    \boldsymbol{\pi}(c, u \mid t, x, i) = \frac{\exp\left( \frac{1}{\lambda} \hat{q}(t, x, i, c, u) \right)}{\int_{\mathcal{A}} \exp\left( \frac{1}{\lambda} \hat{q}(t, x, i, c', u') \right) dc' \, du'},
\end{equation}
assuming the normalizing constant is finite for all $(t, x, i)$, then $\boldsymbol{\pi}$ is the optimal policy and $\hat{J}$ is the corresponding optimal value function.
\end{lemma}

\begin{remark}
Lemma \ref{thm:Martingale Characterization Theorem}-(\romannumeral 1) facilitates \textit{on-policy} learning, where learning the $q$-function of a given target policy $\boldsymbol{\pi}$ is based on the data $$\{(s, X_s^{\boldsymbol{\pi}}, \alpha_s, c_s^{\boldsymbol{\pi}}, u_s^{\boldsymbol{\pi}}), t \leq s \leq T\}$$ generated by $\boldsymbol{\pi}$ itself. On the other hand, Lemma \ref{thm:Martingale Characterization Theorem}-(\romannumeral 2) and -(\romannumeral 3) provide the theoretical foundation for \textit{off-policy} learning, where learning the $q$-function of $\boldsymbol{\pi}$ is based on data generated by a different policy $\boldsymbol{\pi}'$, known as the behavior policy.
\end{remark}

The continuous-time difference (CTD) error differential $d\delta_t$ for the $q$-function is given by the increment of the martingale
\begin{equation}\label{theoretical_TD}
d\delta_t = d\hat{J}(t, X_t, \alpha_t) + \left( U_1(t, c_t) - \hat{q}(t, X_t, \alpha_t; c_t, u_t) \right) dt.
\end{equation}
By the martingale property, the expectation of the stochastic integral of this differential with respect to any predictable test process must be zero. This continuous-time orthogonality condition serves as the foundation for driving the parameter updates in our actor-critic architecture.

\subsection{Algorithm design}\label{sub43}

We adopt a parameterized actor-critic framework that avoids direct integration over continuous action spaces. A key challenge in our regime-switching setting is that the optimal Gibbs policy derived from the entropy-regularized control problem involves a normalizing constant that, in general, cannot be computed in closed form, especially when the action space is jointly continuous over consumption $c$ and investment $u$. Following the approach of \cite{jia2022policy, zhang2026continuous}, we therefore parameterize both the value function and the policy within tractable families of distributions (e.g., Gaussian) and update the network parameters by minimizing the Kullback--Leibler (KL) divergence to the target Gibbs measure, rather than by evaluating the normalizing constant directly.

Let $\phi$ and $\omega$ denote the parameters of the value function $V_\phi(t,x,i)$ and the stochastic policy $\boldsymbol{\pi}_\omega(c,u\mid t,x,i)$, respectively. By It\^{o}'s lemma applied to $V_\phi$ along the grid-sampled trajectory, the $q$-function satisfies
\begin{equation}
\begin{aligned}
q(t_k,X_{t_k},\alpha_{t_k};c_{t_k},u_{t_k})\Delta t=&V(t_{k+1},X_{t_{k+1}},\alpha_{t_{k+1}})-V(t_k,X_{t_k},\alpha_{t_k})\\
&
+ U_1(t_k,c_{t_k})\Delta t + M(t_k,t_{k+1}),
\end{aligned}
\end{equation}
where $M(t_k,t_{k+1})$ is a martingale difference. Dropping it and replacing $V$ by $V_\phi$, the combination $(q-\lambda\ln\boldsymbol{\pi}_\omega)\Delta t$ arising in the policy gradient is approximated by the empirical CTD error
\begin{equation}
\begin{aligned}
\hat{\delta}_{t_k}
=&\, V_\phi(t_{k+1},X_{t_{k+1}},\alpha_{t_{k+1}})-V_\phi(t_k,X_{t_k},\alpha_{t_k})\\
&+\Bigl(U_1(t_k,c_{t_k})-\lambda\ln\boldsymbol{\pi}_\omega(c_{t_k},u_{t_k}\mid t_k,X_{t_k},\alpha_{t_k})\Bigr)\Delta t,
\end{aligned}
\end{equation}
where, at the terminal time $t_K=T$, the value network is naturally anchored by the terminal wealth utility, satisfying $V_\phi(T,X_T,\alpha_T)=U_2(X_T)$.

\subsubsection{Critic update}
The critic parameters $\phi$ are updated to minimize the squared CTD error using a semi-gradient approach. Choosing the test function as $\xi_{t_k} = \partial_{\phi} V_\phi(t_k, X_{t_k}, \alpha_{t_k})$, the update rule is
\begin{equation}
    \phi \leftarrow \phi + \alpha_\phi \partial_{\phi} V_\phi(t_k, X_{t_k}, \alpha_{t_k}) \hat{\delta}_{t_k},
\end{equation}
where $\alpha_\phi > 0$ denotes the learning rate for the critic network.

\subsubsection{Actor update}
To update the actor parameters $\omega$, we minimize the Kullback-Leibler (KL) divergence between the parameterized policy $\boldsymbol{\pi}_\omega$ and the theoretically optimal Gibbs measure. Using the score function $\zeta_{t_k} = \partial_{\omega} \ln \boldsymbol{\pi}_\omega(c_{t_k}, u_{t_k} \mid t_k, X_{t_k}, \alpha_{t_k})$, the model-free update rule driven by the empirical CTD error is
\begin{equation}
    \omega \leftarrow \omega + \alpha_\omega \partial_{\omega} \ln \boldsymbol{\pi}_\omega(c_{t_k}, u_{t_k} \mid t_k, X_{t_k}, \alpha_{t_k}) \hat{\delta}_{t_k},
\end{equation}
where $\alpha_\omega > 0$ represents the learning rate for the actor network.

Below, we summarize the implementation logic in two frameworks: an online incremental algorithm (Algorithm \ref{alg:online_single}) and an offline episodic algorithm (Algorithm \ref{alg:offline_single}). We execute the stochastic policies on a uniform time grid with step size $\Delta t$ such that $\mathbb{S} = \{0 = t_0 < t_1 < \dots < t_K = T\}$.
\begin{algorithm}[ht]
\caption{Online Incremental q-Learning Algorithm}
\label{alg:online_single}
\begin{algorithmic}[1]
\STATE \textbf{Input:} Initial state $x_0 > 0$, initial regime $i_0$, horizon $T$, time step $\Delta t$, number of episodes $N$, number of mesh grids $K$, learning rates $\alpha_\phi, \alpha_\omega$, schedule $l(j)$, and temperature parameter $\lambda$.
\STATE \textbf{Initialize:} Critic parameters $\phi$ and actor parameters $\omega$.
\FOR{\text{episode} $j = 1, 2, \dots, N$}
    \STATE Initialize $k = 0$. Observe initial state $X_{t_0} \leftarrow x_0$ and regime $\alpha_{t_0} \leftarrow i_0$.
    \WHILE{$k < K$}
        \STATE Sample action $(c_{t_k}, u_{t_k}) \sim \boldsymbol{\pi}_\omega(\cdot \mid t_k, X_{t_k}, \alpha_{t_k})$.
        \STATE Apply action $(c_{t_k}, u_{t_k})$ to the \textbf{Environment Simulator}; observe new wealth $X_{t_{k+1}}$, regime $\alpha_{t_{k+1}}$, and utility $r(t_k) = U_1(t_k, c_{t_k})$.
        \STATE Define the next-state evaluation:
         \STATE  $v_{\text{next}} = U_2(X_{t_{k+1}})$ if $k = K-1$; otherwise, $V_\phi(t_{k+1}, X_{t_{k+1}}, \alpha_{t_{k+1}})$.
        \STATE Compute the empirical CTD error:
        \STATE $\hat{\delta}_{t_k} = v_{\text{next}} - V_\phi(t_k, X_{t_k}, \alpha_{t_k}) + \left( r(t_k) - \lambda \ln \boldsymbol{\pi}_\omega(c_{t_k}, u_{t_k} \mid t_k, X_{t_k}, \alpha_{t_k}) \right) \Delta t$
        \STATE Update Critic: 
        \STATE $\phi \leftarrow \phi + l(j) \alpha_\phi \partial_{\phi} V_\phi(t_k, X_{t_k}, \alpha_{t_k}) \hat{\delta}_{t_k}$
        \STATE Update Actor: 
        \STATE $\omega \leftarrow \omega + l(j) \alpha_\omega \partial_{\omega} \ln \boldsymbol{\pi}_\omega(c_{t_k}, u_{t_k} \mid t_k, X_{t_k}, \alpha_{t_k}) \hat{\delta}_{t_k}$
        
        \STATE Update $k \leftarrow k + 1$.
    \ENDWHILE
\ENDFOR
\end{algorithmic}
\end{algorithm}
\begin{algorithm}[ht]
\caption{Offline Episodic q-Learning Algorithm}
\label{alg:offline_single}
\begin{algorithmic}[1]
\STATE \textbf{Input:} Initial state $x_0 > 0$, initial regime $i_0$, horizon $T$, time step $\Delta t$, number of episodes $N$, number of mesh grids $K$, initial learning rates $\alpha_\phi$ and $\alpha_\omega$, schedule $l(j)$, and temperature parameter $\lambda$.
\STATE \textbf{Initialize:} Critic parameters $\phi$ and actor parameters $\omega$.
\FOR{\text{episode} $j = 1, 2, \dots, N$}
    \STATE Initialize $k = 0$. Observe initial state $X_{t_0} \leftarrow x_0$ and regime $\alpha_{t_0} \leftarrow i_0$.
    \WHILE{$k < K$}
        \STATE Sample action $(c_{t_k}, u_{t_k}) \sim \boldsymbol{\pi}_\omega(\cdot \mid t_k, X_{t_k}, \alpha_{t_k})$.
        \STATE Apply action $(c_{t_k}, u_{t_k})$ to the \textbf{Environment Simulator}.
        \STATE Observe new wealth state $X_{t_{k+1}}$, new regime $\alpha_{t_{k+1}}$, and instantaneous utility $r(t_k) = U_1(t_k, c_{t_k})$.
        \STATE Store transition tuple $(t_k, X_{t_k}, \alpha_{t_k}, c_{t_k}, u_{t_k}, r(t_k), X_{t_{k+1}}, \alpha_{t_{k+1}})$ in the episode buffer.
        \STATE Update $k \leftarrow k + 1$.
    \ENDWHILE
    \FOR{\text{step} $k = 0, 1, \dots, K-1$}
        \STATE Define the next-state evaluation:
        \STATE  $v_{\text{next}} = U_2(X_{t_{k+1}})$ if $k = K-1$; otherwise, $V_\phi(t_{k+1}, X_{t_{k+1}}, \alpha_{t_{k+1}})$.
        \STATE Compute the empirical CTD error:
        \STATE $\hat{\delta}_{t_k} = v_{\text{next}} - V_\phi(t_k, X_{t_k}, \alpha_{t_k}) + \left( r(t_k) - \lambda \ln \boldsymbol{\pi}_\omega(c_{t_k}, u_{t_k} \mid t_k, X_{t_k}, \alpha_{t_k}) \right) \Delta t$
    \ENDFOR
    \STATE Compute the cumulative parameter updates:
    \STATE $\Delta \phi = \sum_{k=0}^{K-1} \partial_\phi V_\phi(t_k, X_{t_k}, \alpha_{t_k}) \hat{\delta}_{t_k}$
    \STATE $\Delta \omega = \sum_{k=0}^{K-1} \partial_\omega \ln \boldsymbol{\pi}_\omega(c_{t_k}, u_{t_k} \mid t_k, X_{t_k}, \alpha_{t_k}) \hat{\delta}_{t_k}$
    
    \STATE Apply parameter updates:
    \STATE $\phi \leftarrow \phi + l(j) \alpha_\phi \Delta \phi$
    ; \quad $\omega \leftarrow \omega + l(j) \alpha_\omega \Delta \omega$
\ENDFOR
\end{algorithmic}
\end{algorithm}

\begin{remark}\label{remark:parameterized_distribution}
In practical implementations of the actor update, if the theoretical functional form of the optimal policy distribution is known \textit{a priori}, it is advantageous to parameterize $\boldsymbol{\pi}_\omega$ accordingly. For instance, in our continuous-time consumption-investment optimization problem, the theoretical optimal investment strategy $u$ typically follows a Gaussian distribution, while the optimal consumption $c$ follows a distribution defined on the positive half-line, such as a Gamma distribution. By designing the actor network to directly output the specific statistical parameters of these exact distributions (e.g., the mean and variance for the Gaussian distribution, and the shape and rate parameters for the Gamma distribution), we inject structural theoretical priors into the model. This explicit parameterization significantly reduces the hypothesis space compared to using a generic, black-box density estimator. Consequently, this structure not only accelerates the convergence of the actor-critic algorithm but also enhances training stability.
\end{remark}

\subsection{Numerical examples}
In this subsection, we present numerical experiments to demonstrate the effectiveness of the proposed actor-critic $q$-learning algorithm for the exploratory regime-switching consumption-investment problem. To clarify the experimental setup, we explicitly distinguish between the parameters used to construct the \textbf{Environment simulator} (which governs the underlying dynamics but remains unknown to the learning agent) and the parameters required as known inputs for the learning algorithm.

\textbf{Environment simulator parameters.} The simulator is built upon a financial market with $D=2$ regimes, representing a ``bull'' market (Regime 0) and a ``bear'' market (Regime 1). The hidden market parameters for the two regimes are set as follows.
\begin{itemize}
    \item Regime 0 (Bull market): The risk-free rate $r(0) = 0.03$, expected return $\mu(0) = 0.08$, and volatility $\sigma(0) = 0.15$.
    \item Regime 1 (Bear market): The risk-free rate $r(1) = 0.01$, expected return $\mu(1) = 0.04$, and volatility $\sigma(1) = 0.25$.
\end{itemize}
The generator matrix of the continuous-time Markov chain is given by
\begin{equation}
Q = \begin{pmatrix} -0.5 & 0.5 \\ 1.0 & -1.0 \end{pmatrix},
\end{equation}
which implies that the market tends to recover from the bear state to the bull state faster than it transitions from the bull state to the bear state. 

\textbf{Learning algorithm inputs and settings.} The agent requires several known inputs to execute the learning process. The finite time horizon is set to $T=1.0$, which is divided into $K=50$ mesh grids, yielding a time step of $\Delta t = 0.02$. The temperature parameter is inputted as $\lambda = 0.1$. To evaluate the agent's performance, the objective incorporates both an instantaneous consumption utility function and a terminal utility function. The specific mathematical forms of these utility functions will be explicitly defined in the subsequent subsections for different experimental scenarios.

To approximate the value function, the critic network employs a structured design using a LogSumExp formulation to guarantee monotonicity and  concavity with respect to the wealth variable $x$. The neural networks for the actor and critic parameters ($\omega$ and $\phi$) are trained using the Adam optimizer with initial learning rates $\alpha_\phi = \alpha_\omega = 10^{-3}$ over $N = 820$ episodes.

\subsubsection{Numerical analysis for logarithmic utility function}
We consider the case where both the instantaneous consumption utility and the terminal utility are logarithmic. Specifically, the consumption utility is given by $U_{1}(t,c)=e^{-\rho t}\ln(c)$ with a subjective discount rate $\rho=0.5$, and the terminal utility is defined as $U_2(x) = \ln(x)$. To sample exploratory actions under this setting, the actor network models the investment strategy $u$ using a Gaussian distribution and the consumption strategy $c$ using a Gamma distribution. 
\begin{figure}[htbp]
    \centering
\includegraphics[width=0.8\textwidth]{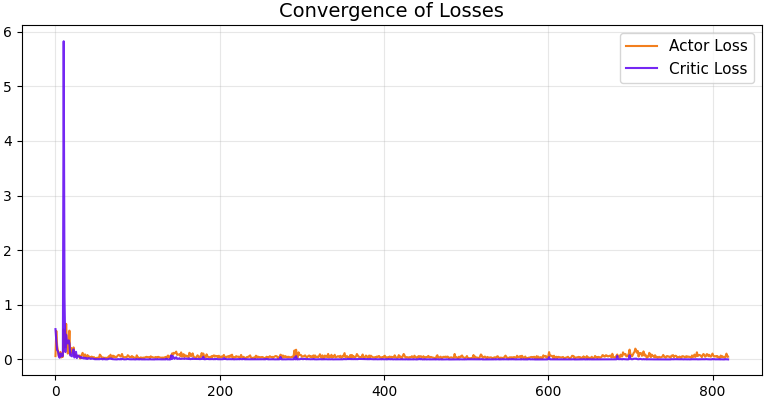}
    \caption{Convergence of Actor and Critic Losses. The figure illustrates a stable loss convergence under logarithmic consumption utility and logarithmic terminal utility.}
    \label{fig:losses}
\end{figure}
\paragraph{Convergence of Losses}
As depicted in Fig. \ref{fig:losses}, the squared CTD error of the Critic network decreases rapidly and stabilizes around zero. Concurrently, the Actor policy loss converges smoothly under the continuous-time policy entropy regularization. This dual stability indicates that the Critic effectively approximates the underlying HJB equation while the Actor maintains a well‑balanced trade‑off between exploration and exploitation.

\begin{figure}[htbp]
    \centering
    \includegraphics[width=0.8\textwidth]{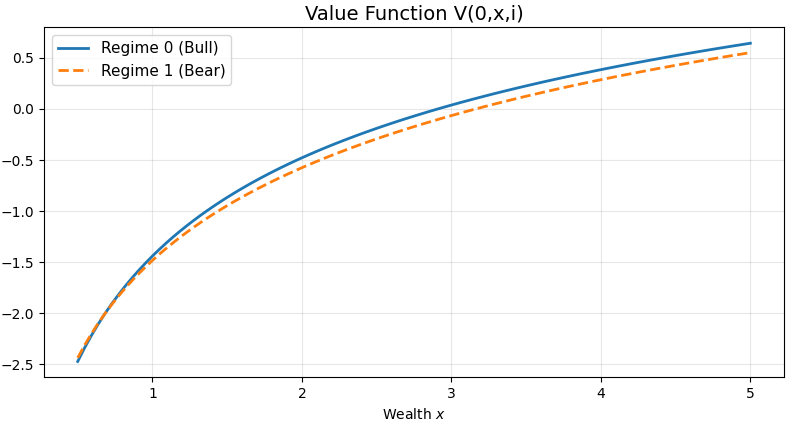}
    \caption{Value Function $V(0,x,i)$. The numerical performance demonstrates monotonic and concave value functions under logarithmic consumption utility and logarithmic terminal utility.}
    \label{fig:value_fn}
\end{figure}
\paragraph{Value Function Properties}
Fig. \ref{fig:value_fn} illustrates the learned value functions $V(0,x,i)$ for Regime 0 (Bull market) and Regime 1 (Bear market). Two economic features emerge as follows.
\begin{itemize}
    \item Monotonicity and concavity: For both market regimes, $V(0,x,i)$ is strictly increasing and strictly concave with respect to wealth $x$, reflecting the investor's risk aversion.
    \item Regime superiority: The value function in Regime 0 consistently dominates that in Regime 1 across the entire wealth spectrum ($V(0,x,0) > V(0,x,1)$), confirming that the policy captures higher expected returns in favorable economic regimes.
\end{itemize}

\begin{figure}[htbp]
    \centering
    \includegraphics[width=0.8\textwidth]{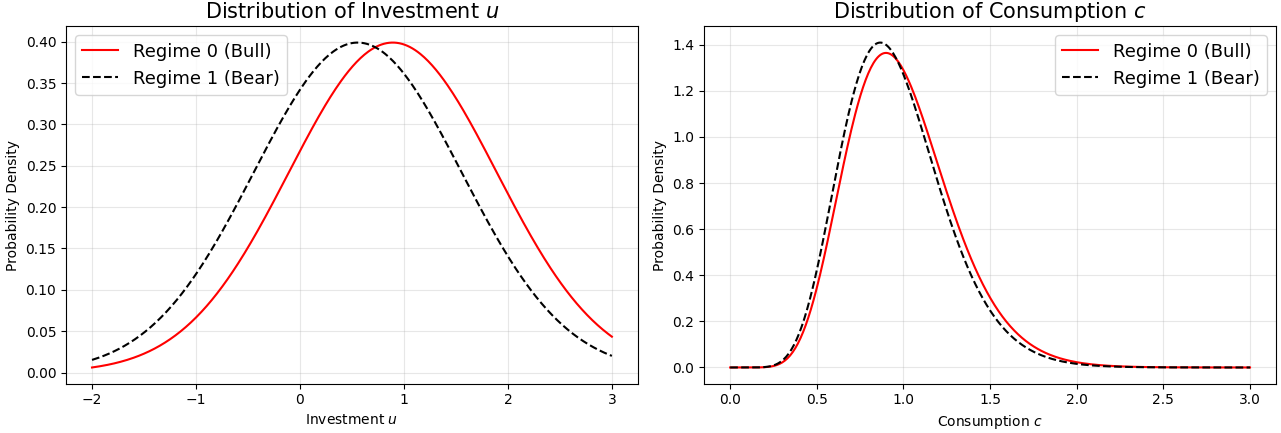}
    \caption{Exploratory Strategy Distributions of $u$ and $c$. The policy parameterization uses a Gaussian distribution for the investment strategy $u$ and a Gamma distribution for the consumption strategy $c$.}
    \label{fig:distributions}
\end{figure}
\paragraph{Optimal Exploratory Strategy Distributions}
Fig. \ref{fig:distributions} illustrates the optimal exploratory policies. To handle continuous action spaces, the Actor network  parameterizes and learns the probability distributions of the actions. Notably, the classical deterministic optimal strategy corresponds to the mode (the peak) of these distributions rather than the mean.
\begin{itemize}
    \item Investment strategy $u$ (Gaussian): Modeled as $u \sim \mathcal{N}(\mu_u, \sigma_u^2)$. Because the Gaussian distribution is symmetric, its mode coincides with its mean $\mu_u$. The peak for Regime 0 (Bull market, red line) is noticeably shifted to the right compared to Regime 1 (Bear market, black dashed line), indicating higher optimal capital allocation and risk exposure during economic prosperity.
   \item Consumption strategy $c$ (Gamma): Modeled as $c \sim \text{Gamma}(\alpha_c, \beta_c)$ to enforce $c > 0$. Due to skewness, the mode distinctly differs from the expected value. Based on the distribution plot, the density peak for Regime 1 (Bear market, black dashed line) is higher and shifted to the left ($c \approx 0.85$) relative to Regime 0 (Bull market, red line, $c \approx 0.90$). This indicates a more concentrated and conservative strategy during the bear regime, restricting immediate consumption, whereas the high-return bull regime permits a slightly higher, more dispersed consumption profile.
\end{itemize}

\subsubsection{Numerical examples for power utility function}
To examine the applicability of the proposed algorithm beyond logarithmic utility, we consider power utility for both consumption and terminal wealth. Specifically, we set $U_1(t,c)=e^{-\rho t}c^{1-\gamma}/(1-\gamma)$ and $U_2(x)=x^{1-\gamma}/(1-\gamma)$, where $\rho=0.5$ and the coefficient of relative risk aversion is $\gamma=0.5$. This is the parameterization discussed in Remark~\ref{rem:power_utility}. 
\begin{figure}[htbp]
    \centering
    \includegraphics[width=0.8\textwidth]{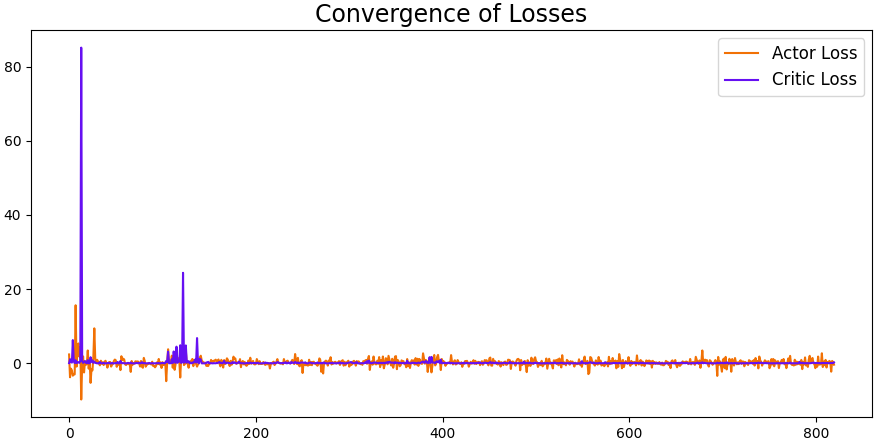}
    \caption{Convergence of Actor and Critic Losses. The figure illustrates stable convergence of the losses under power consumption utility and power terminal utility ($\gamma = 0.5$).}
    \label{fig:losses_power}
\end{figure}
\paragraph{Convergence of Losses}
As illustrated in Fig. \ref{fig:losses_power}, both the Actor and Critic losses demonstrate overall convergence stability across the training episodes. Although a temporary disturbance in the Critic loss occurs around episode $130$, the neural network rapidly self-corrects and stabilizes near zero. This rapid recovery highlights the numerical resilience and algorithmic stability of our continuous-time q-Learning framework in handling non-linear power utility specifications.

\begin{figure}[htbp]
    \centering
    \includegraphics[width=0.8\textwidth]{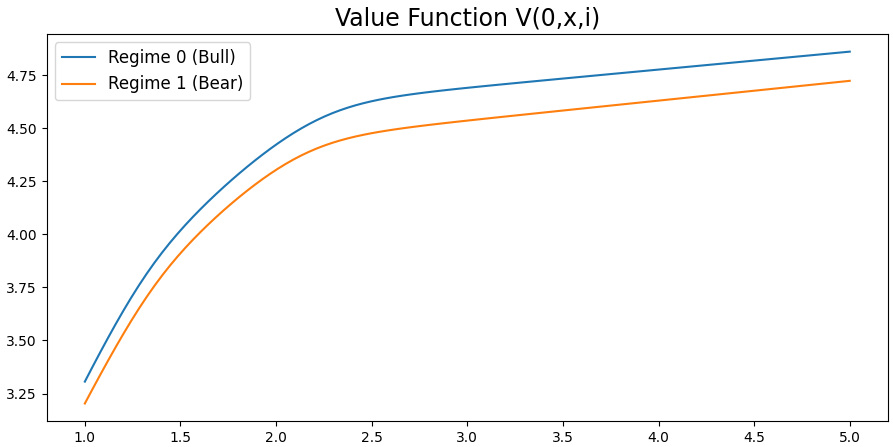}
    \caption{Value Function $V(0,x,i)$. The numerical performance demonstrates monotonic and concave value functions under power consumption utility and power terminal utility ($\gamma = 0.5$).}
    \label{fig:value_fn_power}
\end{figure}
\paragraph{Value Function Properties}
Fig. \ref{fig:value_fn_power} depicts the learned value functions $V(0,x,i)$ at $t=0$ under power utility ($\gamma = 0.5$). The learned values satisfy key economic properties.
\begin{itemize}
    \item Monotonicity and concavity: Across both market regimes, $V(0,x,i)$ remains strictly increasing and concave with respect to wealth $x$, consistent with the investor's persistent risk-averse behavior.
    \item Regime dominance: The value function in Regime 0 (Bull market, blue solid curve)  dominates that in Regime 1 (Bear market, orange dashed curve) over the entire wealth grid ($V(0,x,0) > V(0,x,1)$), corroborating the higher expected reward potential in economic expansions.
\end{itemize}

\begin{figure}[htbp]
    \centering
    \includegraphics[width=0.8\textwidth]{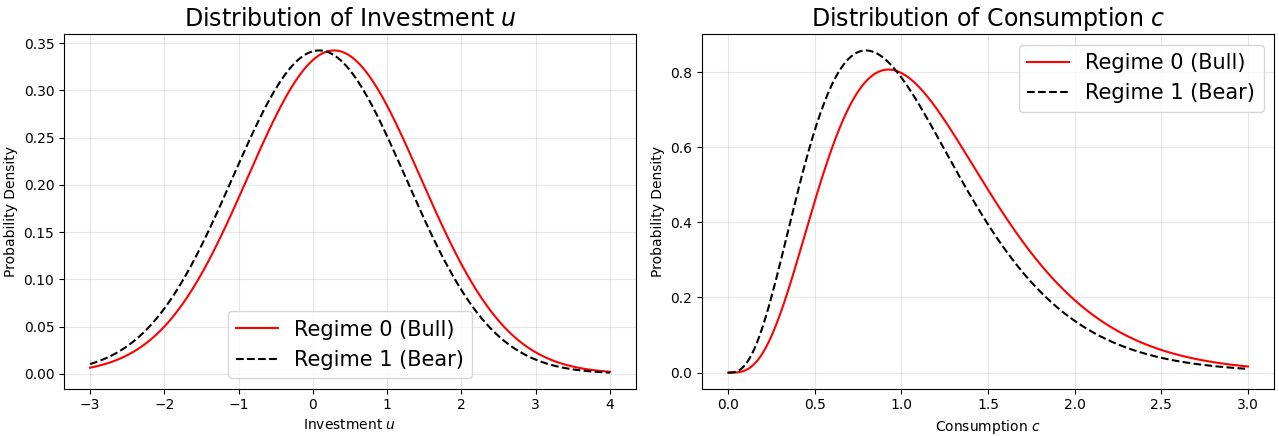}
\caption{Exploratory Strategy Distributions of $u$ and $c$. The policy parameterization uses a Gaussian distribution for the investment strategy $u$, while the empirical distribution of the consumption strategy $c$ shares some structural similarities with a Gamma distribution under power utility ($\gamma = 0.5$).}
\label{fig:distributions_power}
\end{figure}
\paragraph{Optimal Exploratory Strategy Distributions}
As shown in Fig. \ref{fig:distributions_power}, the action probability distributions learned by the Actor network capture state-dependent investment and consumption behaviors.
\begin{itemize}
    \item Investment strategy $u$ (Gaussian): Parameterized as $u \sim \mathcal{N}(\mu_u, \sigma_u^2)$, the mode for Regime 0 (Bull market, red solid curve) is shifted rightward relative to Regime 1 (Bear market, black dashed curve). This demonstrates that the agent dynamically shifts its investment mode to allocate more capital toward the risky asset when market conditions are favorable.
\item Consumption strategy $c$: The learned distribution of $c$  satisfies positivity ($c > 0$) and exhibits an asymmetric profile that is somewhat reminiscent of a Gamma distribution. The density peak for Regime 0 shifts rightward ($\text{mode} \approx 1.0$) compared to Regime 1 ($\text{mode} \approx 0.8$). Unlike the logarithmic utility case, under power utility ($\gamma = 0.5$), the wealth effect induced by favorable regime dynamics dominates, leading the investor to scale up immediate consumption in prosperous market states.
\end{itemize}

\section{Conclusion}
\label{Section 5}
In this paper, we establish an entropy-regularized reinforcement learning (ERRL) framework for the continuous-time consumption-investment problem under a regime-switching diffusion model with general utility functions. By incorporating stochastic exploratory policies and Shannon entropy regularization into the stochastic optimal control formulation, we derive the explicit representations for the optimal exploratory portfolio and consumption strategies. Specifically, we show that the optimal investment strategy follows a multivariate Gaussian distribution, while the optimal consumption policy takes the form of a Gibbs measure, which reduces to a Gamma distribution under logarithmic utility. We also briefly illustrate the corresponding policy structure under power utility. Our analysis reveals that the modes of the optimal exploratory distributions retain the functional forms of the classical deterministic optimal controls, although their values need not coincide at a positive exploration temperature. Furthermore, under logarithmic utility, we establish the convergence of the exploratory value function to its classical unregularized counterpart as the exploration temperature parameter vanishes ($\lambda\to0^+$).

To operate in model-free and unknown market environments, we develop a continuous -time $q$-learning actor-critic algorithm. We apply and specialize the existing theoretical results, proved the policy improvement theorem, and provide a martingale characterization for both on-policy and off-policy learning. Leveraging these theoretical results, we implement an actor-critic parameterization driven by continuous-time temporal difference (CTD) errors. Parameterizing the actor network based on the theoretical distribution families substantially accelerated the algorithm's convergence and enhanced numerical stability. Numerical experiments across regime-switching market environments validated our theoretical findings, demonstrating stable loss convergence, strict monotonicity and concavity of the learned value functions, and consistent regime superiority.

Several promising avenues remain open for future research. First, the proposed framework can be extended to incorporate market frictions such as transaction costs and short-selling constraints, or partially observable regime states. Second, generalizing the current framework to game-theoretic settings (such as stochastic differential games or mean-field games) presents a fascinating direction to study optimal strategies under competition or cooperation. Finally, exploring offline RL and model-based RL methodologies in continuous-time financial control is of great importance for improving sample efficiency and safe exploration in practical financial applications.

\section*{Acknowledgments}
The authors would like to thank the handling editor and
the anonymous reviewers for their insightful comments
and constructive suggestions. The authors are also grateful
to Xin Zhang of Southeast University for valuable advice.


\appendix
\section{Exploratory dynamic system
\eqref{eq:exploratory_sde}}\label{Appendix A}
\setcounter{equation}{0}
\renewcommand{\theequation}{A.\arabic{equation}}

Consider the classical wealth dynamic system
\cref{classical SDE}. We derive the corresponding
exploratory wealth dynamics using the averaged
infinitesimal generator associated with a
randomized policy.
For a fixed state
$(s,x,i)\in[0,T]\times(0,\infty)\times\mathcal{M}$
and a deterministic action $(c,u)\in\mathcal{A}$,
the drift and instantaneous variance of the
classical wealth process are given by
\begin{equation*}
\begin{aligned}
b(s,x,i;c,u)
&=
x\big[
r(s,i)+u^\top\sigma(s,i)\theta(s,i)
\big]-c,\\
a(s,x,i;c,u)
&=
x^2u^\top\sigma(s,i)\sigma^\top(s,i)u.
\end{aligned}
\end{equation*}
For any sufficiently smooth test function
$\varphi(s,x,i)$, the infinitesimal generator
of the regime-switching wealth process under
the fixed action $(c,u)$ is
\begin{equation*}
\mathcal{L}^{c,u}\varphi(s,x,i)
=
b(s,x,i;c,u)\varphi_x(s,x,i)+\frac12 a(s,x,i;c,u)
\varphi_{xx}(s,x,i)+\sum_{j=1}^{D}q_{ij}\varphi(s,x,j).
\end{equation*}
Under the randomized policy
$\boldsymbol{\pi}(c,u\mid s,x,i)$,
the exploratory wealth dynamics are
characterized by the averaged infinitesimal
generator
\begin{equation*}
\begin{aligned}
\mathcal{L}^{\boldsymbol{\pi}}
\varphi(s,x,i)
=&
\int_{\mathcal{A}}
\mathcal{L}^{c,u}\varphi(s,x,i)
\boldsymbol{\pi}(c,u\mid s,x,i)
\,dc\,du\\
={}&
b^{\boldsymbol{\pi}}(s,x,i)
\varphi_x(s,x,i)+\frac12
a^{\boldsymbol{\pi}}(s,x,i)
\varphi_{xx}(s,x,i)+\sum_{j=1}^{D}
q_{ij}\varphi(s,x,j),
\end{aligned}
\end{equation*}
where the averaged drift and instantaneous
variance are respectively defined by
\begin{equation*}
\begin{aligned}
b^{\boldsymbol{\pi}}(s,x,i)
=&
\int_{\mathcal{A}}
\Big\{
x\big[
r(s,i)+u^\top\sigma(s,i)\theta(s,i)
\big]-c
\Big\}
\boldsymbol{\pi}(c,u\mid s,x,i)
\,dc\,du,\\
a^{\boldsymbol{\pi}}(s,x,i)
=&
\int_{\mathcal{A}}
x^2u^\top\sigma(s,i)\sigma^\top(s,i)u
\boldsymbol{\pi}(c,u\mid s,x,i)
\,dc\,du.
\end{aligned}
\end{equation*}

The averaging of the generator corresponds
to averaging the drift and quadratic
variation characteristics of the classical
wealth process under the randomized policy.
Consequently, the exploratory wealth process
can be represented by 
\begin{equation*}
\left\{\begin{array}{ll}
dX(s) = &\displaystyle \bigg( \int_{\mathcal{A}} \Big\{ X(s) \big[ r(s, \alpha(s)) + u^\top \sigma(s, \alpha(s))\theta(s, \alpha(s)) \big] - c \Big\} \\
&\quad\quad\times\boldsymbol{\pi}(c, u \mid s, X(s), \alpha(s)) \,dc\,du \bigg) ds \\
&\displaystyle + \bigg\{ \int_{\mathcal{A}} X(s)^2 u^\top \sigma(s, \alpha(s))\sigma^\top(s, \alpha(s)) u\\
&\quad\quad\times\, \boldsymbol{\pi}(c, u \mid s, X(s), \alpha(s)) \,dc\,du \bigg\}^{1/2} d\tilde{W}(s), \\
X(t) = &x > 0, \quad \alpha(t) = i.
\end{array} \right.
\end{equation*}
Here, $\tilde W(s)$ is a standard
one-dimensional Brownian motion. The regime-switching
component remains governed by the
generator matrix $Q$.

\section{Proof of Proposition \ref{pro:Well-posedness of the exploratory SDE}}\label{Appendix B}
\setcounter{equation}{0}
\renewcommand{\theequation}{B.\arabic{equation}}
\begin{proof}
For notational simplicity, we denote the drift and diffusion coefficients of the SDE
\eqref{eq:exploratory_sde} under policy $\boldsymbol{\pi}$ as
$b_{\boldsymbol{\pi}}(s,x,i)$ and $\sigma_{\boldsymbol{\pi}}(s,x,i)$, respectively
\begin{equation*}
\begin{aligned}
b_{\boldsymbol{\pi}}(s, x, i) =& x \left( r(s,i) + \int_{\mathcal{A}} u^\top \sigma(s,i) \theta(s,i) \boldsymbol{\pi}(c,u\mid s,x,i)\,dc\,du \right)\\
&- \int_{\mathcal{A}} c \boldsymbol{\pi}(c,u\mid s,x,i)\,dc\,du, \\[1mm]
\sigma_{\boldsymbol{\pi}}(s, x, i) =& \left\{ x^2 \int_{\mathcal{A}} u^\top \sigma(s,i) \sigma^\top(s,i) u \boldsymbol{\pi}(c,u\mid s,x,i)\,dc\,du \right\}^{1/2}.
\end{aligned}
\end{equation*}

\noindent\textbf{Local Lipschitz continuity.}
Fix a compact interval $[\varepsilon,K]\subset(0,\infty)$ and let
$x,y\in[\varepsilon,K]$. $C_{\varepsilon,K}>0$ denotes a constant that may change from line to line. By condition (ii) of Definition
\ref{def:admissible_strategy}, together with the boundedness of
$\sigma$ and $\theta$, we have
\begin{equation*}
\begin{aligned}
&\left|
\int_{\mathcal A}u^\top\sigma(s,i)\theta(s,i)
\left[
\boldsymbol{\pi}(c,u\mid s,x,i)
-\boldsymbol{\pi}(c,u\mid s,y,i)
\right]dc\,du
\right|\\
\leq&
\|\sigma\theta\|_\infty
\int_{\mathcal A}|u|\,
\left|\boldsymbol{\pi}(c,u\mid s,x,i)
-\boldsymbol{\pi}(c,u\mid s,y,i)\right|dc\,du\\
\leq&
\|\sigma\theta\|_\infty
\int_{\mathcal A}(1+u^\top u)\,
\left|\boldsymbol{\pi}(x)-\boldsymbol{\pi}(y)\right|dc\,du
\leq
C_{\varepsilon,K}|x-y|,
\end{aligned}
\end{equation*}
where we used $|u|\leq 1+u^\top u$, and $\|\sigma\theta\|_\infty:=\sup_{(t,j)\in[0,T]\times\mathcal{M}}\|\sigma(t,j)\theta(t,j)\|$ is the uniform supremum norm. Similarly,
\[
\left|
\int_{\mathcal A}c
\left[
\boldsymbol{\pi}(c,u\mid s,x,i)
-\boldsymbol{\pi}(c,u\mid s,y,i)
\right]dc\,du
\right|
\leq C_{\varepsilon,K}|x-y|.
\]
Moreover, by condition (iv) and the Cauchy--Schwarz inequality,
\[
\left|
\int_{\mathcal A}u^\top\sigma(s,i)\theta(s,i)
\boldsymbol{\pi}(c,u\mid s,x,i)\,dc\,du
\right|
\leq
\|\sigma\theta\|_\infty
\left(\int_{\mathcal A}u^\top u\,\boldsymbol{\pi}\,dc\,du\right)^{1/2}
\leq C.
\]
Therefore,
\[
\begin{aligned}
&|b_{\boldsymbol{\pi}}(s,x,i)-b_{\boldsymbol{\pi}}(s,y,i)|\\
=&
\biggl|
x\Bigl(r(s,i)+\int_{\mathcal A}u^\top\sigma(s,i)\theta(s,i)\boldsymbol{\pi}(c,u\mid s,x,i)\,dc\,du\Bigr)\\
&
-y\Bigl(r(s,i)+\int_{\mathcal A}u^\top\sigma(s,i)\theta(s,i)\boldsymbol{\pi}(c,u\mid s,y,i)\,dc\,du\Bigr)\\
&
-\Bigl(\int_{\mathcal A}c\,\boldsymbol{\pi}(c,u\mid s,x,i)\,dc\,du
-\int_{\mathcal A}c\,\boldsymbol{\pi}(c,u\mid s,y,i)\,dc\,du\Bigr)
\biggr|\\
\leq&
|x-y|\,\Bigl|r(s,i)+\int_{\mathcal A}u^\top\sigma(s,i)\theta(s,i)\boldsymbol{\pi}(c,u\mid s,x,i)\,dc\,du\Bigr|\\
&
+y\,\Bigl|\int_{\mathcal A}u^\top\sigma(s,i)\theta(s,i)
\bigl[\boldsymbol{\pi}(c,u\mid s,x,i)-\boldsymbol{\pi}(c,u\mid s,y,i)\bigr]\,dc\,du\Bigr|\\
&
+\Bigl|\int_{\mathcal A}c\,
\bigl[\boldsymbol{\pi}(c,u\mid s,x,i)-\boldsymbol{\pi}(c,u\mid s,y,i)\bigr]\,dc\,du\Bigr|\leq C_{\varepsilon,K}|x-y|.
\end{aligned}
\]
Hence $b_{\boldsymbol{\pi}}(s,x,i)$ is locally Lipschitz continuous in $x$.

Next, for the diffusion coefficient, set
\[
a_{\boldsymbol{\pi}}(s,x,i)
:=
\int_{\mathcal A}u^\top\sigma(s,i)\sigma^\top(s,i)u\,
\boldsymbol{\pi}(c,u\mid s,x,i)\,dc\,du.
\]
Since $\sigma$ is uniformly bounded, condition (ii) gives
\begin{equation*}
\begin{aligned}
|a_{\boldsymbol{\pi}}(s,x,i)-a_{\boldsymbol{\pi}}(s,y,i)|
&\leq
\|\sigma\sigma^\top\|_\infty\int_{\mathcal A}u^\top u
\left|
\boldsymbol{\pi}(c,u\mid s,x,i)
-\boldsymbol{\pi}(c,u\mid s,y,i)
\right|dc\,du\\
&\leq C_{\varepsilon,K}|x-y|.
\end{aligned}
\end{equation*}
Furthermore, by
$\sigma(s,i)\sigma^\top(s,i)\geq\delta I$ and condition (iv),
\[
a_{\boldsymbol{\pi}}(s,x,i)
\geq
\delta\int_{\mathcal A}u^\top u\,
\boldsymbol{\pi}(c,u\mid s,x,i)\,dc\,du
\geq\delta\underline{\epsilon}>0.
\]
On the other hand, condition (iv) and the boundedness of $\sigma$ imply
that $a_{\boldsymbol{\pi}}(s,x,i)\leq\overline C$ for some constant
$\overline C>0$. Therefore,
\begin{equation*}
\begin{aligned}
\left|
\sigma_{\boldsymbol{\pi}}(s,x,i)
-\sigma_{\boldsymbol{\pi}}(s,y,i)
\right|
&=
\left|
x\sqrt{a_{\boldsymbol{\pi}}(s,x,i)}
-y\sqrt{a_{\boldsymbol{\pi}}(s,y,i)}
\right|\\
&\leq
\sqrt{\overline C}\,|x-y|
+
y\left|
\sqrt{a_{\boldsymbol{\pi}}(s,x,i)}
-\sqrt{a_{\boldsymbol{\pi}}(s,y,i)}
\right|.
\end{aligned}
\end{equation*}
Using the identity
\[
\left|\sqrt{a_{\boldsymbol{\pi}}(s,x,i)}
-\sqrt{a_{\boldsymbol{\pi}}(s,y,i)}\right|
=
\frac{
|a_{\boldsymbol{\pi}}(s,x,i)-a_{\boldsymbol{\pi}}(s,y,i)|
}{
\sqrt{a_{\boldsymbol{\pi}}(s,x,i)}
+\sqrt{a_{\boldsymbol{\pi}}(s,y,i)}
}
\leq
\frac{C_{\varepsilon,K}}
{2\sqrt{\delta\underline{\epsilon}}}|x-y|,
\]
and $y\leq K$, it follows that
\[
\left|
\sigma_{\boldsymbol{\pi}}(s,x,i)
-\sigma_{\boldsymbol{\pi}}(s,y,i)
\right|
\leq
\left(
\sqrt{\overline C}
+
\frac{K C_{\varepsilon,K}}
{2\sqrt{\delta\underline{\epsilon}}}
\right)|x-y|.
\]
Thus, $\sigma_{\boldsymbol{\pi}}(s,x,i)$ is locally Lipschitz continuous
in $x$ on $(0,\infty)$.

\medskip
\noindent\textbf{Linear growth condition.}
Since $r$, $\sigma$, and $\theta$ are uniformly bounded, there exist constants
$M_r,M_\sigma,M_\theta>0$ such that
$|r(s,i)|\leq M_r$,
$\|\sigma(s,i)\|\leq M_\sigma$,
$\|\theta(s,i)\|\leq M_\theta$.
By the definition of $b_{\boldsymbol{\pi}}$ and the triangle inequality,
\[
\begin{aligned}
|b_{\boldsymbol{\pi}}(s,x,i)|
\leq&
x|r(s,i)|
+x\left|
\int_{\mathcal A}u^\top\sigma(s,i)\theta(s,i)
\boldsymbol{\pi}(c,u\mid s,x,i)\,dc\,du
\right|\\
&
+
\int_{\mathcal A}c\,\boldsymbol{\pi}(c,u\mid s,x,i)\,dc\,du.
\end{aligned}
\]
By the Cauchy--Schwarz inequality and condition (iv),
\[
\begin{aligned}
\left|
\int_{\mathcal A}u^\top\sigma(s,i)\theta(s,i)
\boldsymbol{\pi}(c,u\mid s,x,i)\,dc\,du
\right|
&\leq
M_\sigma M_\theta
\int_{\mathcal A}|u|\,
\boldsymbol{\pi}(c,u\mid s,x,i)\,dc\,du\\
&\leq
M_\sigma M_\theta
\left(
\int_{\mathcal A}u^\top u\,
\boldsymbol{\pi}(c,u\mid s,x,i)\,dc\,du
\right)^{1/2}\\
&\leq M_\sigma M_\theta\sqrt{C}.
\end{aligned}
\]
Therefore, using again condition (iv),
\[
\begin{aligned}
|b_{\boldsymbol{\pi}}(s,x,i)|
&\leq
\left(
M_r+M_\sigma M_\theta\sqrt C+C
\right)x
\leq K_1(1+x)
\end{aligned}
\]
for some constant $K_1>0$.
Similarly,
\[
\begin{aligned}
|\sigma_{\boldsymbol{\pi}}(s,x,i)|
&=
x\left\{
\int_{\mathcal A}
u^\top\sigma(s,i)\sigma^\top(s,i)u\,
\boldsymbol{\pi}(c,u\mid s,x,i)\,dc\,du
\right\}^{1/2}\\
&\leq
x\left\{
M_\sigma^2
\int_{\mathcal A}u^\top u\,
\boldsymbol{\pi}(c,u\mid s,x,i)\,dc\,du
\right\}^{1/2}
\leq M_\sigma\sqrt C\,x
\leq K_1(1+x).
\end{aligned}
\]
Hence, both $b_{\boldsymbol{\pi}}$ and $\sigma_{\boldsymbol{\pi}}$
satisfy the linear growth condition.

By the standard existence and uniqueness theorem for SDEs with locally
Lipschitz coefficients and linear growth, the exploratory SDE
\eqref{eq:exploratory_sde} admits a unique maximal strong solution on $[t,\tau_0)$, where
\[
\tau_0:=\inf\{s\ge t:X(s)=0\}
\]
is the first hitting time of zero. 

\medskip
\noindent\textbf{Strict positivity.}
It remains to show that $\tau_0>T$ almost surely. For $n\ge1$ define
\[
\tau_n:=\inf\{s\ge t:X(s)\le 1/n\}.
\]
For all sufficiently large $n$, $1/n<x$, so $X(s)>0$ on $[t,\tau_n]$. On this interval, Itô's formula applied to $\ln X(s)$ gives
\[
d\ln X(s)=\widetilde\mu(s)\,ds+\widetilde\nu(s)\,d\widetilde W(s),
\]
where
\[
\widetilde\mu(s):=\frac{b_{\boldsymbol{\pi}}(s,X(s),\alpha(s))}{X(s)}-\frac12 a_{\boldsymbol{\pi}}(s,X(s),\alpha(s)),
\qquad
\widetilde\nu(s):=\sqrt{a_{\boldsymbol{\pi}}(s,X(s),\alpha(s))}.
\]
Since
\[
\frac{b_{\boldsymbol{\pi}}(s,x,i)}{x}
=r(s,i)+\int u^\top\sigma\theta\,\boldsymbol{\pi}(\cdot\mid s,x,i)\,dc\,du-\frac1x\int c\,\boldsymbol{\pi}(\cdot\mid s,x,i)\,dc\,du,
\]
condition (iv) and Cauchy--Schwarz yield $|b_{\boldsymbol{\pi}}(s,x,i)/x|\le M_r+M_\sigma M_\theta\sqrt C+C=:M_b$. Together with $a_{\boldsymbol{\pi}}\le\overline C$, this gives
\[
|\widetilde\mu(s)|\le M_b+\tfrac12\overline C=:M',
\qquad
|\widetilde\nu(s)|\le\sqrt{\overline C}
\qquad\text{a.s. on }[t,\tau_n].
\]
Integrating on $[t,s\wedge\tau_n]$,
\[
\ln X(s\wedge\tau_n)=\ln x+\int_t^{s\wedge\tau_n}\widetilde\mu(r)\,dr+M_n(s),
\qquad
M_n(s):=\int_t^{s\wedge\tau_n}\widetilde\nu(r)\,d\widetilde W(r).
\]
Here $M_n$ is a continuous square-integrable martingale with
\[
\langle M_n\rangle_s=\int_t^{s\wedge\tau_n}a_{\boldsymbol{\pi}}(r,X(r),\alpha(r))\,dr\le\overline C(T-t),
\]
so Doob's $L^2$ inequality gives the $n$-uniform bound
\[
\mathbb E\Bigl[\sup_{s\in[t,T]}|M_n(s)|^2\Bigr]\le 4\,\mathbb E[\langle M_n\rangle_T]\le 4\overline C(T-t).
\]
On the event $\{\tau_n\le T\}$, continuity gives $X(\tau_n)=1/n$, hence
\[
-\ln n=\ln x+\int_t^{\tau_n}\widetilde\mu(r)\,dr+M_n(\tau_n).
\]
Since $|\int_t^{\tau_n}\widetilde\mu\,dr|\le M'(T-t)$, we get
$M_n(\tau_n)\le-\ln n-\ln x+M'(T-t)$. Thus, for $L_n:=\ln n+\ln x-M'(T-t)>0$,
\[
\{\tau_n\le T\}\subseteq\Bigl\{\sup_{s\in[t,T]}|M_n(s)|\ge L_n\Bigr\}.
\]
By Chebyshev's inequality,
\[
\mathbb P(\tau_n\le T)
\le\frac{\mathbb E[\sup_{s\in[t,T]}|M_n(s)|^2]}{L_n^2}
\le\frac{4\overline C(T-t)}{\bigl[\ln n+\ln x-M'(T-t)\bigr]^2}\longrightarrow 0.
\]
Since $\tau_n\uparrow\tau_0$ (by continuity of $X$), the event $\{\tau_0\le T\}$ equals $\bigcap_{n\ge n_0}\{\tau_n\le T\}$ up to finitely many initial indices, so continuity of probability yields
\[
\mathbb P(\tau_0\le T)=\lim_{n\to\infty}\mathbb P(\tau_n\le T)=0.
\]
Hence $\mathbb P(X(s)>0,\ \forall s\in[t,T])=1$. 

Finally, the linear growth condition excludes finite-time explosion, while the preceding argument excludes hitting the boundary $0$. Hence the maximal strong solution extends uniquely to the whole interval $[t,T]$.
\end{proof}

\section{Proof of Lemma \ref{lemma:Monotonicity and concavity}}\label{Appendix C}
\setcounter{equation}{0}
\renewcommand{\theequation}{C.\arabic{equation}}
\begin{proof}
Fix $(t,x,i)$. By the entropy variational formula, finiteness of the
right-hand side of \eqref{eq:HJB_exploratory} requires
\[
    \int_{\mathcal A}
    \exp\left(
        \frac{1}{\lambda}
        H(t,x,i,v_x,v_{xx};c,u)
    \right)\,dc\,du<\infty.
\]
Using \eqref{eq:classical_hamiltonian}, the action-dependent part of the Hamiltonian separates as
\[
    U_1(t,c)-cv_x
    +
    \frac12 x^2v_{xx}u^\top
    \sigma\sigma^\top u
    +
    xv_xu^\top\sigma\theta.
\]

First, if $v_x<0$, then $-cv_x\to\infty$ linearly as $c\to\infty$.
If $v_x=0$, since $U_1(t,\cdot)$ is increasing,
\[
    \int_0^\infty
    \exp\left(\frac{U_1(t,c)}{\lambda}\right)dc
    =\infty.
\]
Hence $v_x>0$.

Next, by the nondegeneracy condition
$\sigma\sigma^\top\geq\delta I$, if $v_{xx}>0$, the quadratic term in $u$
is positive and dominates the linear term as $\lVert u\rVert\to\infty$,
so the integral over $\mathbb R^N$ is infinite. If $v_{xx}=0$, the
integrand is either exponential in a nonzero linear function of $u$, or
constant in $u$ when $v_x\sigma\theta=0$; in either case its integral over
$\mathbb R^N$ is infinite. Therefore $v_{xx}<0$.
Thus $v$ is strictly increasing and strictly concave in $x$.
\end{proof}

\section{Proof of Proposition \ref{Thm_Value_Function_Convergence}}\label{Appendix D}
\setcounter{equation}{0}
\renewcommand{\theequation}{D.\arabic{equation}}

\begin{proof}
Define
\[
A_0(t)=1+\int_t^T e^{-\rho s}\,ds,
\quad
A_\lambda(t)=A_0(t)+\lambda(T-t).
\]
Under logarithmic utility, the classical and exploratory
HJB equations admit solutions of the form
\[
V^{cl}(t,x,i)=A_0(t)\ln x+B_0(t,i),\quad
V(t,x,i)=A_\lambda(t)\ln x+B_\lambda(t,i).
\]
Substituting these expressions into the corresponding
HJB equations yields
\begin{equation}\label{eq:D_B_system}
-\partial_t B_k(t,i)
=F_k(t,i)+\sum_{j=1}^D q_{ij}B_k(t,j),
\qquad B_k(T,i)=0,
\end{equation}
for $k\in\{0,\lambda\}$, where
\begin{equation*}
F_0(t,i)
=A_0(t)\left(r(t,i)
+\frac12\theta(t,i)^\top\theta(t,i)\right)+e^{-\rho t}
\left(\ln\frac{e^{-\rho t}}{A_0(t)}-1\right),
\end{equation*}
and
\begin{align*}
F_\lambda(t,i)
={}&A_\lambda(t)\left(r(t,i)
+\frac12\theta(t,i)^\top\theta(t,i)\right)+\frac{\lambda N}{2}
\ln\frac{2\pi\lambda}{A_\lambda(t)}\\
&
-\frac{\lambda}{2}
\ln\det\bigl(\sigma(t,i)\sigma^\top(t,i)\bigr)+\lambda\ln\Gamma\left(
1+\frac{e^{-\rho t}}{\lambda}\right)+(e^{-\rho t}+\lambda)
\ln\frac{\lambda}{A_\lambda(t)}.
\end{align*}
Since $A_\lambda(t)\ge1$, the boundedness and
nondegeneracy assumptions on the market coefficients
ensure that the associated optimal Gaussian and Gamma
policies are admissible. Standard verification arguments,
together with the required martingale integrability
conditions, identify these solutions with the respective
value functions.

Next, by Stirling's formula, uniformly in $t\in[0,T]$,
\begin{align*}
&\lambda\ln\Gamma\left(
1+\frac{e^{-\rho t}}{\lambda}\right)
+(e^{-\rho t}+\lambda)
\ln\frac{\lambda}{A_\lambda(t)}\\
&\quad=e^{-\rho t}
\left(\ln\frac{e^{-\rho t}}{A_\lambda(t)}-1\right)
+\frac{\lambda}{2}\ln\lambda
+\mathcal O(\lambda).
\end{align*}
Since $A_\lambda(t)-A_0(t)=\mathcal O(\lambda)$
uniformly in $t$, it follows that
\begin{equation}\label{eq:D_error}
F_\lambda(t,i)-F_0(t,i)
=\frac{N+1}{2}\lambda\ln\lambda
+\mathcal O(\lambda),
\end{equation}
uniformly in $(t,i)\in[0,T]\times\mathcal M$.

Writing $B_k(t)=(B_k(t,1),\ldots,B_k(t,D))^\top$,
the solution of \eqref{eq:D_B_system} satisfies
\[
B_k(t)=\int_t^T e^{Q(s-t)}F_k(s)\,ds.
\]
Since $Q$ is a Markov chain generator,
$e^{Q(s-t)}$ is a stochastic matrix. Therefore,
\eqref{eq:D_error} implies
\[
B_\lambda(t,i)-B_0(t,i)
=\frac{N+1}{2}(T-t)\lambda\ln\lambda
+\mathcal O(\lambda),
\]
uniformly in $(t,i)$.
Consequently,
\begin{equation*}
V(t,x,i)-V^{cl}(t,x,i)
=\lambda(T-t)\ln x+\frac{N+1}{2}(T-t)\lambda\ln\lambda
+\mathcal O(\lambda).
\end{equation*}
For any compact interval
$[\varepsilon,R]\subset(0,\infty)$,
the function $\ln x$ is uniformly bounded.
Hence, for sufficiently small $\lambda>0$,
\[
\sup_{\substack{(t,x,i)\in
[0,T]\times[\varepsilon,R]\times\mathcal M}}
|V(t,x,i)-V^{cl}(t,x,i)|
\le C_{\varepsilon,R}\lambda|\ln\lambda|.
\]
Since $\lambda|\ln\lambda|\to0$ as
$\lambda\to0^+$, the desired locally uniform
convergence follows.
\end{proof}
\end{document}